\RequirePackage{fix-cm}%
\documentclass[smallextended]{svjour3}       
\smartqed  

\usepackage{hyperref}

\usepackage{graphicx}
\usepackage{subcaption}
\usepackage{multirow}
\usepackage{units}
\usepackage{hyperref}
\usepackage{listings}
\usepackage{color}
\usepackage{textcomp}
\usepackage{academicons}
\usepackage[dvipsnames,table,xcdraw]{xcolor}

\definecolor{orcidlogocol}{HTML}{A6CE39}

\newcommand{\blackcell}{\cellcolor[HTML]{000000}}
\newcommand{\greycell}[1]{\cellcolor[HTML]{EFEFEF}{#1}}
\newcommand{\greencell}{\cellcolor[HTML]{B9FCAA}}
\newcommand{\dgreencell}{\cellcolor[HTML]{68E54C}}
\newcommand{\ddgreencell}{\cellcolor[HTML]{22A700}}
\newcommand{\dddgreencell}{\cellcolor[HTML]{156800}}
\newcommand{\redcell}{\cellcolor[HTML]{FFA0A0}}
\newcommand{\dredcell}{\cellcolor[HTML]{E90000}}
\newcommand{\ddredcell}{\cellcolor[HTML]{950202}}
\newcommand{\dddredcell}{\cellcolor[HTML]{680000}}
\newcommand{\greenredcell}{\cellcolor[HTML]{706E4B}}
\newcolumntype{C}[1]{>{\centering\let\newline\\\arraybackslash\hspace{0pt}}m{#1}}

\begin{document}

\title{On Multi-Human Multi-Robot Remote Interaction
}
\subtitle{A Study of Transparency, Inter-Human Communication, and Information Loss in Remote Interaction}


\author{Jayam Patel   \and
        Prajankya Sonar \and
        Carlo Pinciroli
}


\institute{J. Patel \at
              Worcester Polytechnic Institute \\
              \email{jupatel@wpi.com}
           \and
           P. Sonar \at
              Worcester Polytechnic Institute \\
              \email{prajankya@wpi.com}
           \and
           C. Pinciroli \at
              Worcester Polytechnic Institute \\
              \email{cpinciroli@wpi.com}
}

\date{Received: \hspace{2cm} / Accepted: \hspace{2cm}}

\maketitle


\begin{abstract}
  In this paper, we investigate how to design an effective interface for remote multi-human multi-robot interaction. While significant research exists on interfaces for individual human operators, little research exists for the multi-human case. Yet, this is a critical problem to solve to make complex, large-scale missions achievable in which direct human involvement is impossible or undesirable, and robot swarms act as a semi-autonomous agents. This paper’s contribution is twofold. The first contribution is an exploration of the design space of computer-based interfaces for multi-human multi-robot operations. In particular, we focus on \emph{information transparency} and on the factors that affect \emph{inter-human communication} in ideal conditions, i.e., without communication issues. Our second contribution concerns the same problem, but considering increasing degrees of \emph{information loss}, defined as intermittent reception of data with noticeable gaps between individual receipts. We derived a set of design recommendations based on two user studies involving 48 participants.
\end{abstract}

\section{Introduction}

Robot swarms promise solutions for missions in which direct human involvement is either impossible or undesirable, such as search-and-rescue, firefighting, planetary exploration, and ocean restoration~\cite{murphy2012decade}. When robot swarms are deployed to perform complex missions, autonomy is only part of the picture. Along with autonomy, it is equally important for human operators to monitor and affect the behavior of the swarm. This creates the issue of designing effective solutions for remote interaction between humans and robot swarms.

While a significant body of work exists in remote interaction involving single humans and one or more robots, the scenario in which \emph{multiple} humans interact with a robot swarm has received little attention. In this paper, we argue that it will be common for multiple humans to cooperate in the supervision of robot swarms. First, the amount of information generated by robot swarms is likely to exceed the span of apprehension of any individual operator~\cite{miller1956magical}, even when considering highly skilled ones such as video gamers. Cooperation among human operators would make monitoring more efficient. Second, the involvement of multiple humans allows for improved flexibility in robot control and task assignment, an important advantage in complex operations.

However, the involvement of multiple humans comes with old and new challenges. Among the old, we highlight the need for \emph{information transparency}, which is the ability of the interface-swarm system to convey useful data for the operators to understand and modify the status of the swarm~\cite{wohleber_effects_2017,chen_situation_2018,roundtree_transparency:_2019,bhaskara_agent_2020,chakraborti_explicability_nodate,tulli_eects_nodate}. Multiple operators also create the new challenge of conveying intentions and actions to other operators, i.e., effective \emph{inter-human communication}, for better cooperation and conflict mitigation~\cite{tomasello2010origins}. Inter-human communication can be either \emph{direct} or \emph{indirect}. Direct communication includes verbal and non-verbal communication (e.g., gestures)~\cite{holdcroft1976forms}. Indirect communication is mediated through the remote interface (e.g., a graphical user interface on a laptop or tablet). Effective indirect communication requires \emph{inter-operator transparency}, which pushes for interface designs that make it simple for operators far away from each other to exchange information on their intentions and plans~\cite{breazeal2005effects,lyons2013being,chen_situation_2014,wohleber_effects_2017,roundtree_transparency:_2019,bhaskara_agent_2020}.

In this paper, we explore the design space of remote interfaces for multi-human multi-robot interaction. We study the role of direct and indirect communication among operators, and investigate how to achieve high levels of information and inter-operator transparency through several variants of our interface. The result of this work is a set of recommendations on which design elements contribute to making a remote interface effective. This part of our study builds upon previous work \cite{Patel2021} in which we investigated transparency and inter-human communication on the performance of human operators in \emph{proximal interaction}. Proximal interaction occurs when humans and robots share the same environment.

Remote interaction allows us to study another important aspect—the role of \emph{information loss}. In this paper, we consider information loss as a decrease in the frequency of the visual information presented to the operators. We measure information loss as the time interval, measured in seconds, between the delivery of consecutive video frames (the inverse of frames per second). Packet loss, bandwidth limitations, and geographical distance between the locations of the operators and the robots act as causal factors for information loss. Information loss leads to degraded operator performance, lack of awareness and trust, and increase in cognitive workload~\cite{ellis2004generalizeability}.

The last factor we consider in our study is that, in presence of non-ideal communication, it is also likely that the operators experience \emph{heterogeneous} levels of information loss, causing a disparity in workload and situational awareness across operators.

The main contributions of this paper can be summarized as follows:
\begin{itemize}
\item We provide an extensive investigation of the design space of remote interfaces for multi-human multi-robot interaction. We consider factors such as direct and indirect communication, information and inter-operator transparency, and homogeneous and heterogeneous information loss.
\item We compile a set of design recommendations validated through a user study that included 48 participants. We implemented a highly configurable remote interface that incorporates these recommendations and enables future studies of this kind.
\end{itemize}

This paper is organized as follows. We discuss related literature on remote human-robot interaction in Sec.~\ref{sec-loi:background}. In Sec.~\ref{sec-loi:interface}, we discuss the design of our configurable remote interface. We report the results of our user study in ideal conditions in Sec.~\ref{sec-loi:tcstudy}. We then introduce different types of information loss and report the results of a dedicated user study in Sec.~\ref{sec-loi:informationloss}. We summarize our contributions and outline directions for future work in Sec.~\ref{sec:conclusion}.

\section{Related Work}
\label{sec-loi:background}

Remote robot control and manipulations has been a field of interest since Goertz and Thompson laid the foundation of modern tele-operation~\cite{goertz1954electronically}. The field has mostly focused manipulators~\cite{hokayem_bilateral_2006,lichiardopol_survey_2007,varkonyi_survey_2014,jung_robotic_2018,li_operator_2018} rather than on mobile robots. This body of research has contributed advancements in tele-presence~\cite{ferreira_immersive_nodate,klow_privacy_2017,dimitoglou_telepresence_2019,salichs_privacy_2019}, tele-robotics~\cite{nak_young_chong_remote_2000,rakita_remote_2019}, tele-operation~\cite{jingtai_liu_competitive_2005,ma_teleoperation_2010,hutchison_evaluation_2010,mansour_dynamic_2012,hong_visual_2013}, and tele-surgery \cite{patel_long_2019,shahzad_telesurgery_2019,dardona_remote_2019}. This research has focused on identifying suitable interfaces and improving their usability~\cite{lager_remote_2019,lunghi_multimodal_2019,roldan_bringing_2019,regenbrecht_intuitive_nodate,music_humanrobot_2019,esfahlani_mixed_2019,welburn_mixed_2019,jang_omnipotent_2019}, as well as proposing novel control architectures for these interfaces~\cite{cheung_semi-autonomous_2011,do_multiple_2011,lee_development_2012,lee_semiautonomous_2013}. Chen \emph{et al.} \cite{chen_human_2007} categorize existing research according to the factors that affect remote control of robots. These factors are field of view, system orientation, camera viewpoints, depth perception, degraded video quality, time delay, and camera motion. Building upon this work, Feth \emph{et al.}~\cite{dillmann_shared-control_2009} and Kim \emph{et al.}~\cite{kim_implementation_2012,kim_implementation_2013} present a shared control framework to allow multiple operators to interact with manipulators. Lee \emph{et al.}~\cite{dong_gun_lee_human-centered_2013} extend these shared control frameworks to study the impact of information delay on the performance of human operators. In their work, the authors incorporate a passivity-based controller to counteract the negative effects of information delay on operator's performance. These works are limited to interface design for remote interaction with industrial manipulators, and their findings may not be applicable to remote interface for manipulating numerous mobile robots. To the best of my knowledge, our study is the first study that investigates the impact of transparency and inter-human communication on a multi-human multi-robot interaction. 

Loss of information has been recognized as a key factor in the performance and engagement of human operators~\cite{chen_human_2007,mackenzie1993lag,ellis2004generalizeability,lane2002effects,sheridan1963remote,rastogi1997design,darken2014spatial,watson1998effects,massimino1994teleoperator,chen2008human}. Research suggests that the effect of information loss and the ability to handle the loss may vary according to the tasks and the interface to interact with the system. To overcome the degradation in performance, there are three methods to mitigate the effects of loss on the performance of human operators. These methods are adopting passivity-based control methods~\cite{lewis_two_2011,varkonyi_survey_2014,cheung_semi-autonomous_2011,jung_robotic_2018,hokayem_bilateral_2006}, predictive displays~\cite{keskinpala2004objective,baker2004improved,calhoun200611,collett2006developer,daily2003world,nielsen2006comparing,sheridan2002humans,kheddar2014virtual,ricks2004ecological} and higher-granularity of control~\cite{patel2019,ayanian_controlling_2014,kolling_human_2013}. However, these studies are limited to the scenario in which a single operator interacts with one of more robots. Our study furthers this line of research by providing an extensive investigation of the factors that affect the design of remote interfaces for \emph{multi}-human multi-robot interaction in presence of information loss.

\section{System Design}
\label{sec-loi:interface}
\begin{figure}[t]
    \centering
    \includegraphics[width=\textwidth]{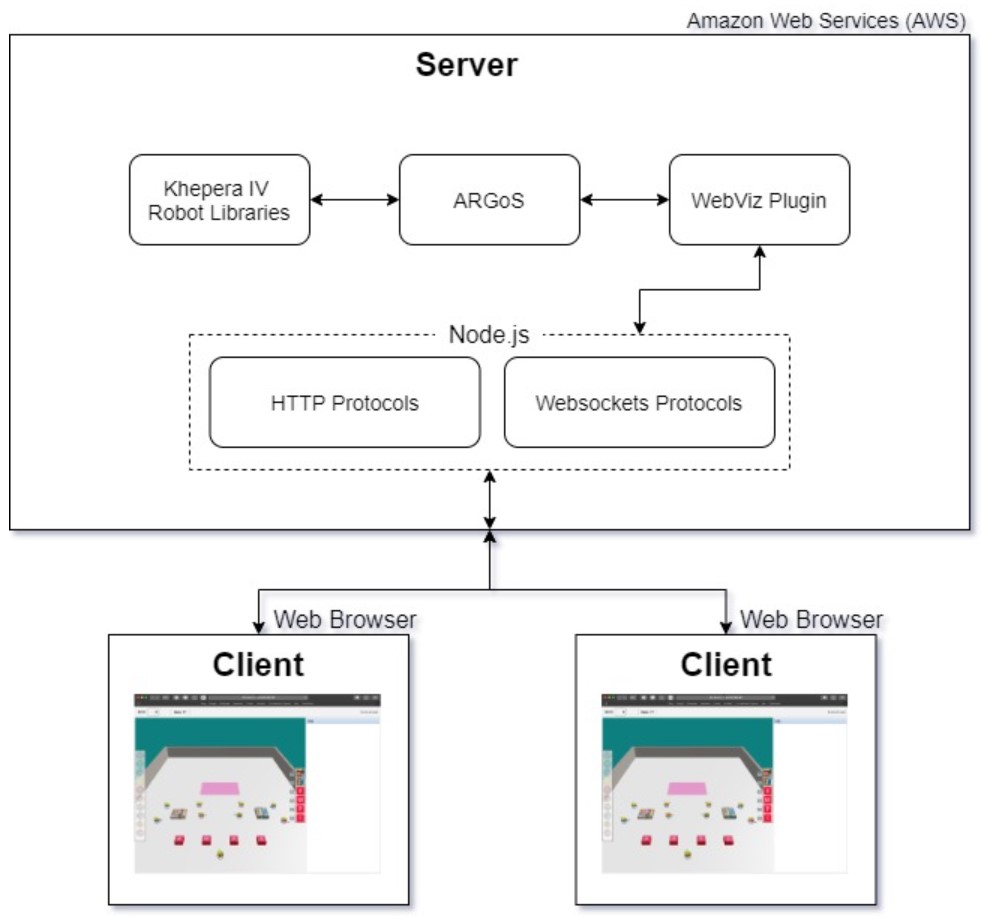}
    \caption{System overview.}
    \label{fig-loi:system_overview}
\end{figure}

\begin{figure}[t]
    \centering
    \includegraphics[width=\textwidth]{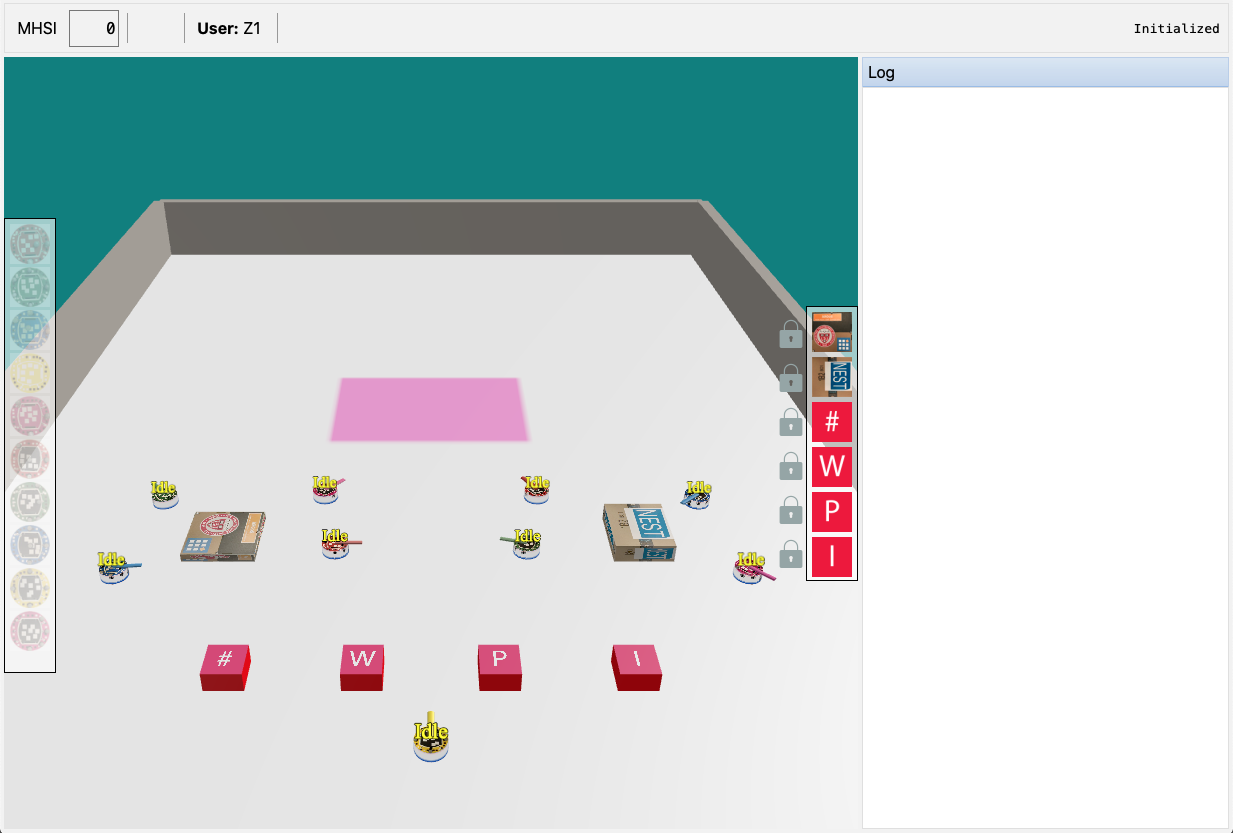}
    \caption{Screenshot of the interface running on an internet browser.}
    \label{fig-loi:screenshot}
\end{figure}

In this section, we present the main features of our remote interface and the behavior of the robots. At its essence, our interface is a web-based client-server architecture. The server runs ARGoS~\cite{Pinciroli:SI2012}, a fast multi-robot simulator, on a node offered by Amazon Web Services\footnote{\url{https://aws.amazon.com/}}. The server is implemented as a visualization plugin that accepts multiple connections from the clients. The client side is a web application implemented with Node.js\footnote{\url{https://nodejs.org/}} and WebGL\footnote{\url{https://get.webgl.org/}} which offers similar features with respect to the original graphical visualization of ARGoS. A diagram of the client-server architecture is reported in Fig.~\ref{fig-loi:system_overview} and a screenshot of the web interface is shown in Fig.~\ref{fig-loi:screenshot}. The source code of the system is available online as open source software.\footnote{\url{https://github.com/NESTLab/argos3-webviz}}

The process starts when a user performs a command on the client. The web interface allows the user to operate at multiple levels of granularity. In our previous work~\cite{patel_mixed-granularity_2019}, we found that mixed granularity of control offers superior usability in complex missions that require both navigation and environment modification. Similarly to~\cite{patel_mixed-granularity_2019}, in this paper we focus on a collective transport scenario due to the compositional nature that this kind of task presents — collective transport combines navigation, task allocation, and object manipulation. Our interface is therefore designed for this scenario and it mirrors many of the features we presented in~\cite{patel_mixed-granularity_2019}. It is important to highlight, however, that the remote interface presented here is a completely new artifact based on a different technology: in fact, the work in~\cite{patel_mixed-granularity_2019} studied \emph{proximal} interactions with a \emph{touch-based} interface.

\subsection{Collective Transport}
\begin{figure}[t]
    \centering
    \includegraphics[width=0.9\textwidth]{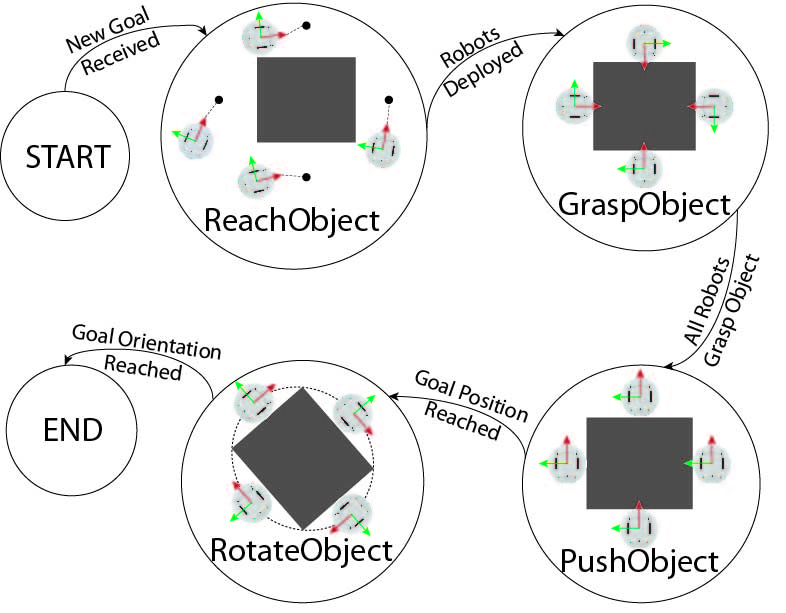}
    \caption{Collective transport state machine.}
    \label{fig-loi:collective_transport}
\end{figure}

We employ a collective transport behavior based on the finite state machine shown in Fig.~\ref{fig-loi:collective_transport}. The behavior is identical to the one discussed in our previous work~\cite{patel_mixed-granularity_2019}. The states in the finite state machine are as follows:

\textbf{Reach Object.} On receiving the desired goal position for the object, the robots in the transport team navigate and organize themselves around the object in a circular manner. These positions are generated based on the number of robots in the team and their distance from the object. The state comes to an end once all the robots reach their designated positions.

\textbf{Approach Object.} After organizing themselves, the robots move towards the centroid of the object. The state comes to an end once all the robots are touching the object.

\textbf{Push Object.} Once the robots are in contact with the object, the robots rotate in place to face the direction of the goal. The robots start moving at equal speed towards the goal, while maintaining a 
fixed distance from the centroid of the object. This strategy prevents the robot in front and on the sides from breaking formation. If a robot breaks the formation, the robots switch back to Reach Object, wait for its completion, and subsequently resume their transport operation. The state comes to an end once the object reaches the goal position.

\textbf{Rotate Object.} The robots rearrange themselves around the object and move in a circular path in the
outward direction, thereby rotating the object in place. If any robot breaks the formation, the robots rearrange themselves and resume rotating the object. The state comes to an end once the robots achieve the desired rotation.

\subsection{User Interface}
\begin{figure}[t]
  \centering
  \begin{subfigure}{0.49\textwidth}
    \includegraphics[width=\textwidth]{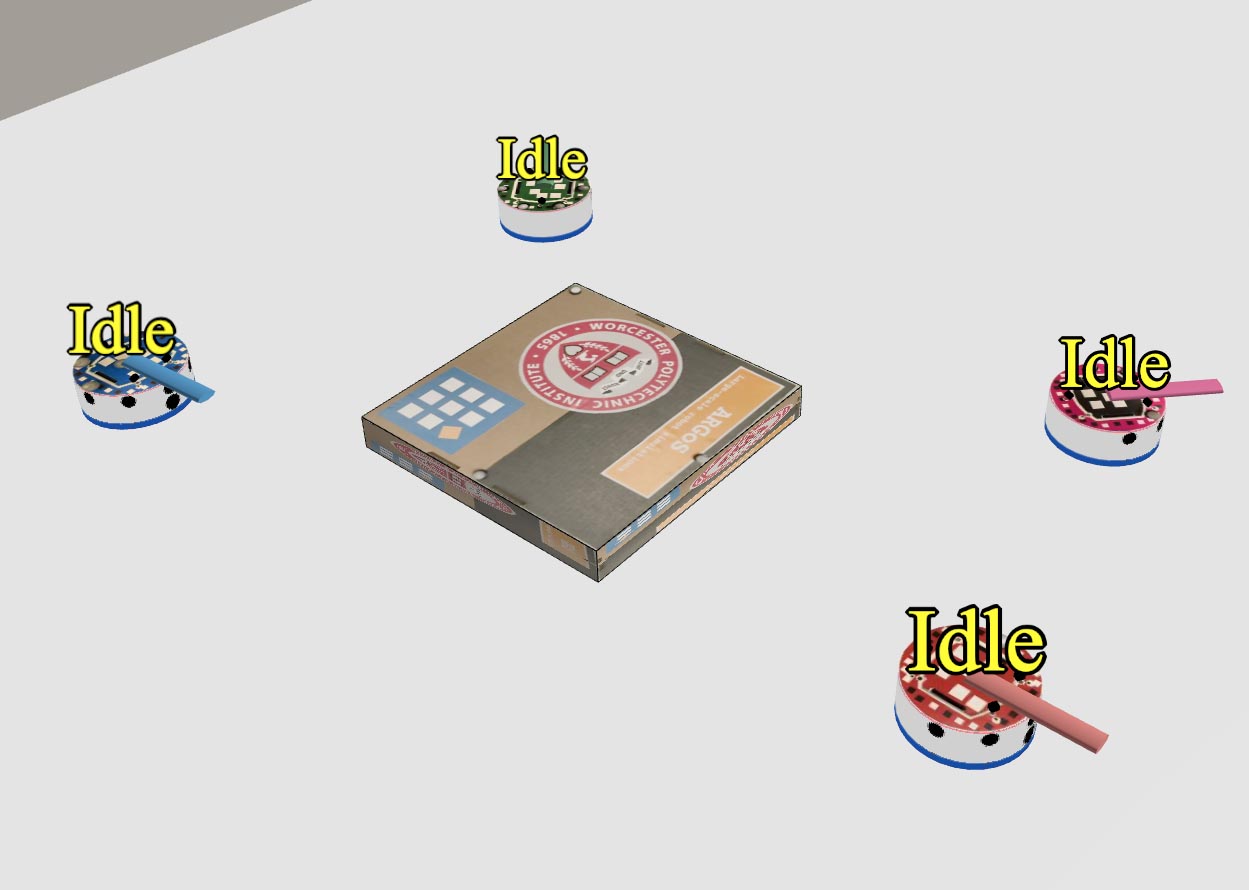}
    \caption{Object recognition}
    \label{fig-loi:modeO1}
  \end{subfigure}
  \begin{subfigure}{0.49\textwidth}
    \includegraphics[width=\textwidth]{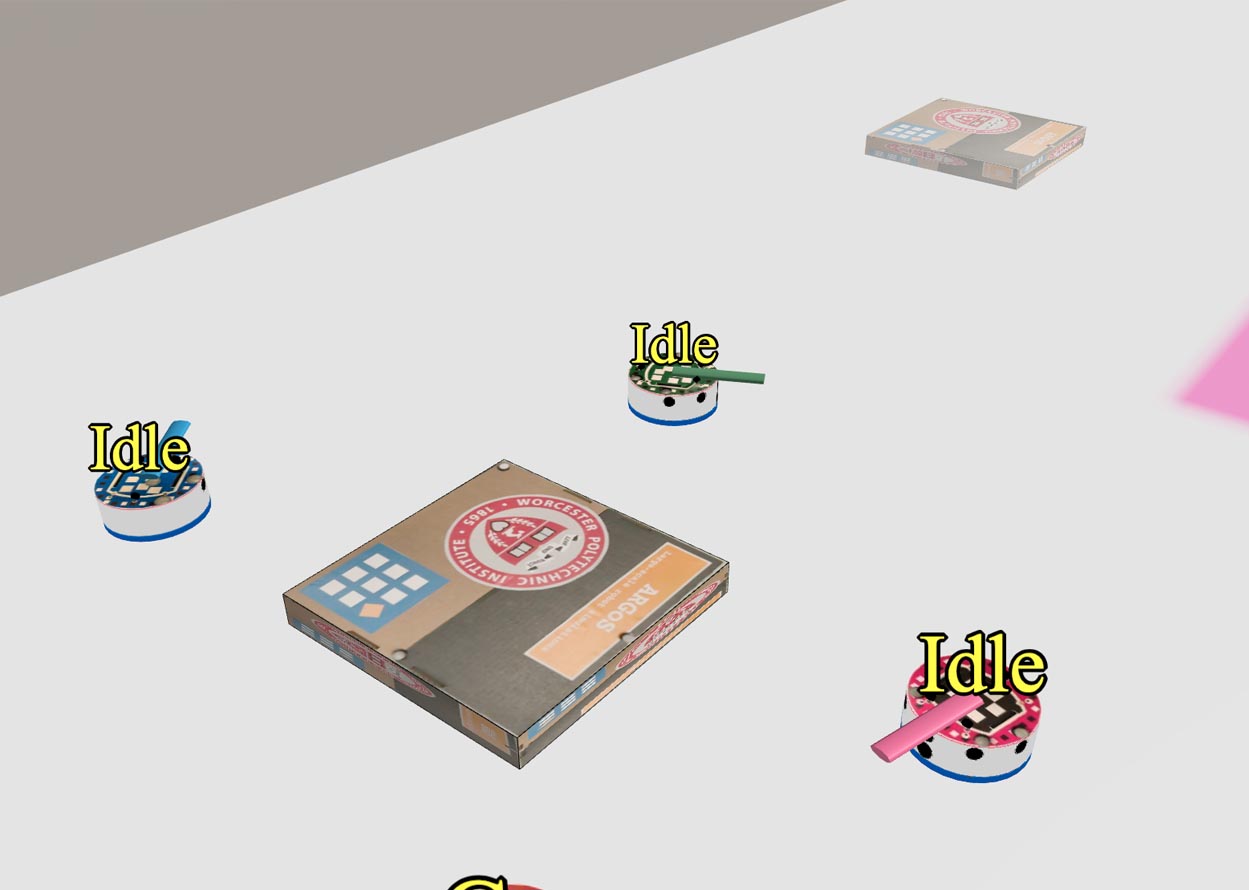}
    \caption{New Goal Defined}
    \label{fig-loi:modeO2}
  \end{subfigure}
  \begin{subfigure}{0.49\textwidth}
    \includegraphics[width=\textwidth]{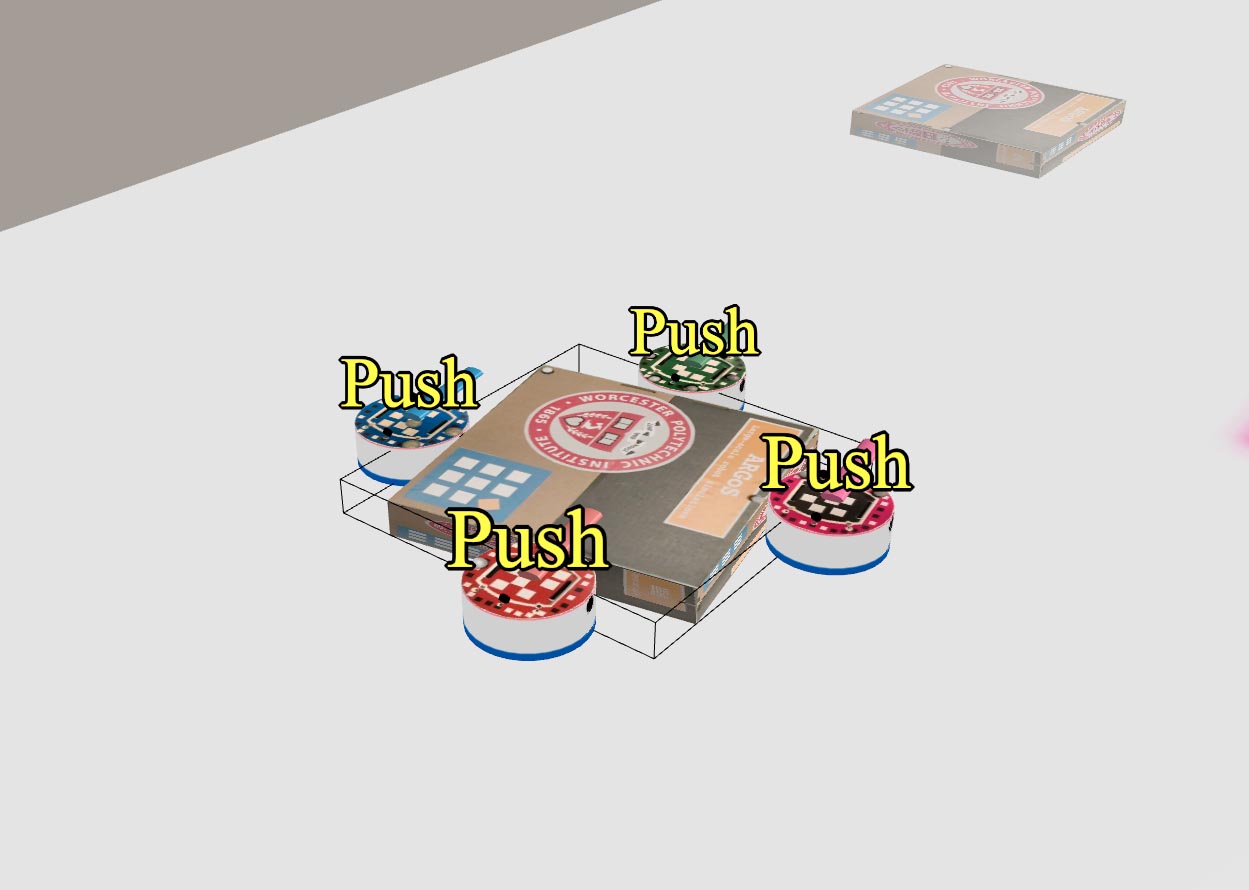}
    \caption{Robots push the object}
    \label{fig-loi:modeO3}
  \end{subfigure}
  \begin{subfigure}{0.49\textwidth}
    \includegraphics[width=\textwidth]{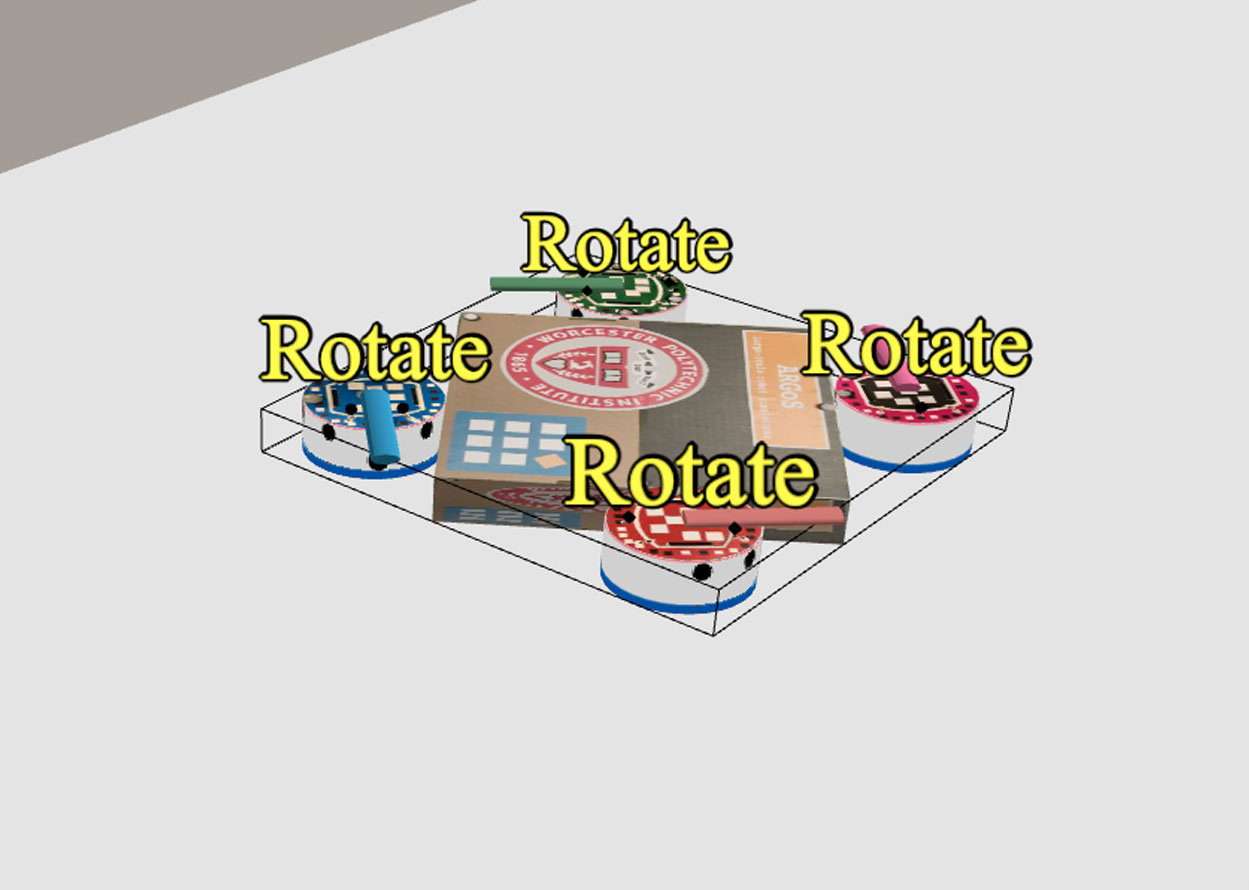}
    \caption{Robots rotate the object}
    \label{fig-loi:modeO4}
  \end{subfigure}
  \caption{Object manipulation by interaction with the object through
    the interface.}\label{fig-loi:modeO}
\end{figure}

\textbf{Object Manipulation.} Object manipulation is triggered when an operator selects an object with a left click. The goal position always requires a right click, and the interface overlays the selected object with a transparent bounding box. The operator can also define the goal position for multiple objects. In this case, the robots autonomously distributed across the objects and transport them using the collective transport behavior. If two or more operators manipulate the same object, the interface keeps the position specified by the last operator. Fig.~\ref{fig-loi:modeO1} shows a selected object overlaid with a bounding box. Fig.~\ref{fig-loi:modeO2} illustrates how the goal position is visualized. The desired position and orientation of the object is conveyed by the interface as shown in Fig.~\ref{fig-loi:modeO3} and~\ref{fig-loi:modeO4}.

\begin{figure}[t]
  \centering
  \begin{subfigure}[t]{0.49\textwidth}
    \includegraphics[width=\textwidth]{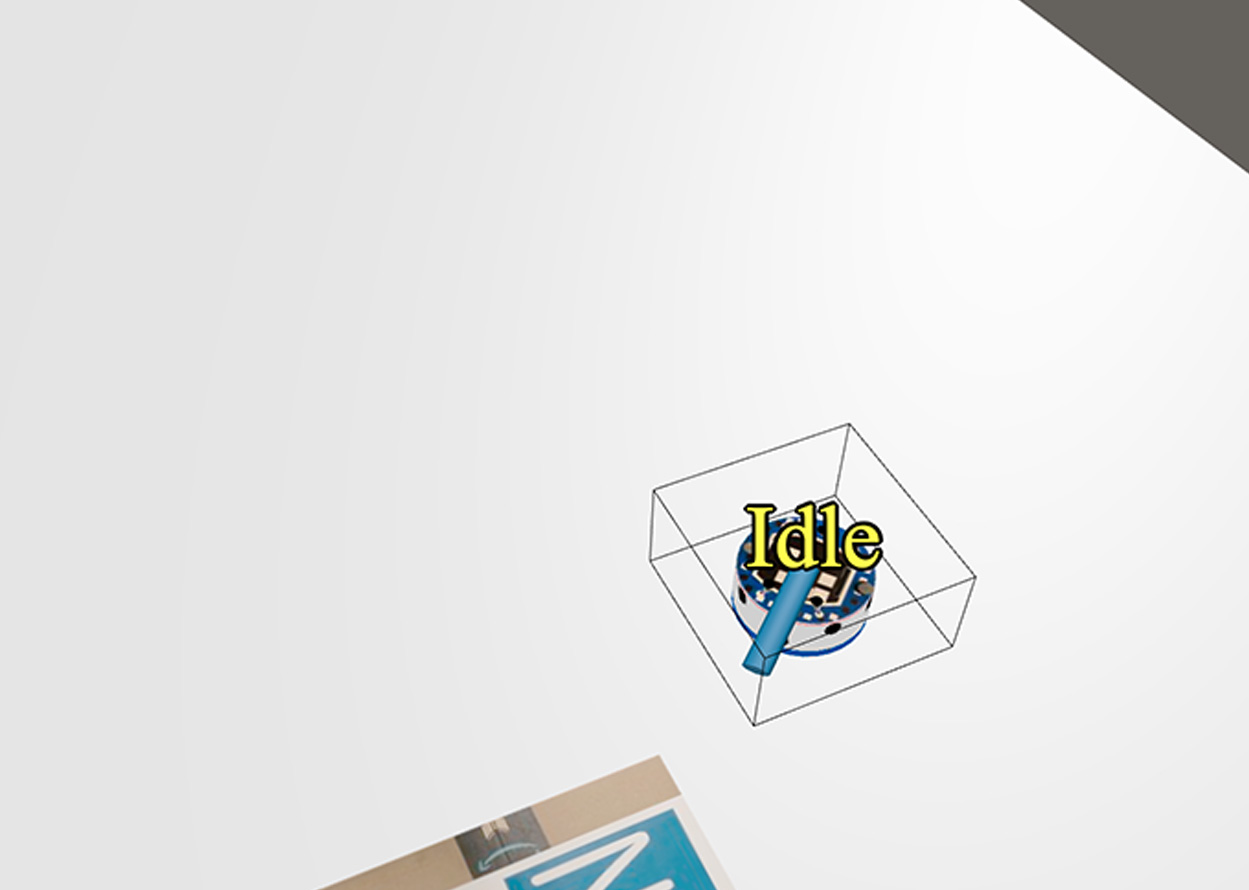}
    \caption{Robot selection}
    \label{fig-loi:modeR1}
  \end{subfigure}
  \begin{subfigure}[t]{0.49\textwidth}
    \includegraphics[width=\textwidth]{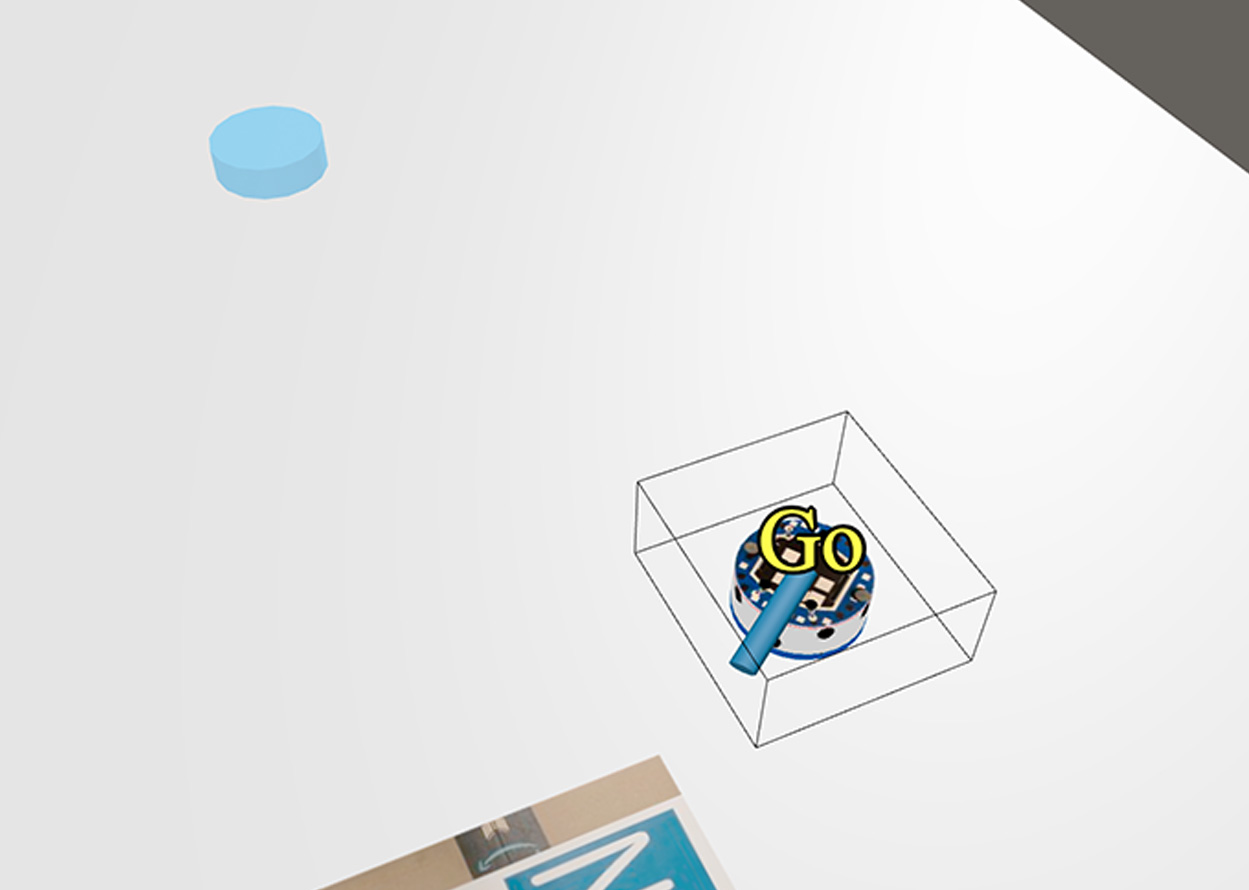}
    \caption{New robot position}
    \label{fig-loi:modeR2}
  \end{subfigure}
  \begin{subfigure}[t]{0.49\textwidth}
    \includegraphics[width=\textwidth]{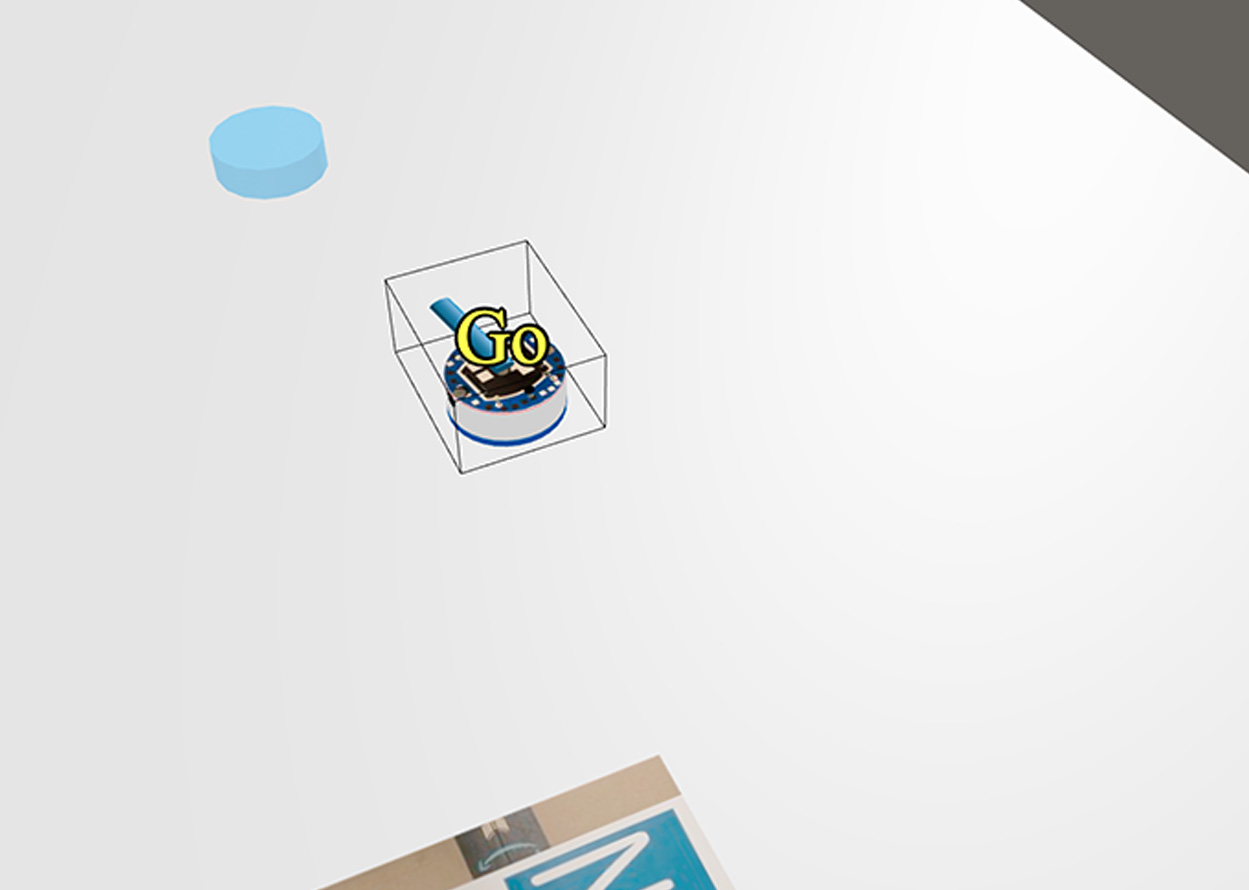}
    \caption{Robot navigating to new position}
    \label{fig-loi:modeR3}
  \end{subfigure}
  \caption{Robot manipulation by interacting with the robots through the
    interface.}\label{fig-loi:modeR}
\end{figure}

\textbf{Robot Manipulation.} Robot manipulation starts with an operator selecting a robot with a left click. The goal position is assigned using a right click. The interface overlays the selected robot with a transparent bounding box convey the current selection. The operator can define the goal position for multiple robots at once. If the robot is performing the collective transport behavior during this request, other robots in the collective transport team pause their operation until the selected robot reaches the desired position. In case the robot is a part of an operator-defined team, the selected robot navigates to the newly specified position and other robots continue their respective operations. When two or more operators want to manipulate the same robot, the interface processes the position specified by the last operator. Fig.~\ref{fig-loi:modeR1} shows a selected robot overlaid with a bounding box to visualize the current selection. Fig.~\ref{fig-loi:modeR2} shows the goal position determined by the operator and visualized as a colored representation of the selected robot. The color of the goal position matches the color of the fiducial markers to differentiate between the goal positions of different robots. Fig.~\ref{fig-loi:modeR3} shows the selected robot navigating to the specified goal position.

\begin{figure}[t]
  \centering
  \begin{subfigure}{0.49\textwidth}
    \includegraphics[width=\textwidth]{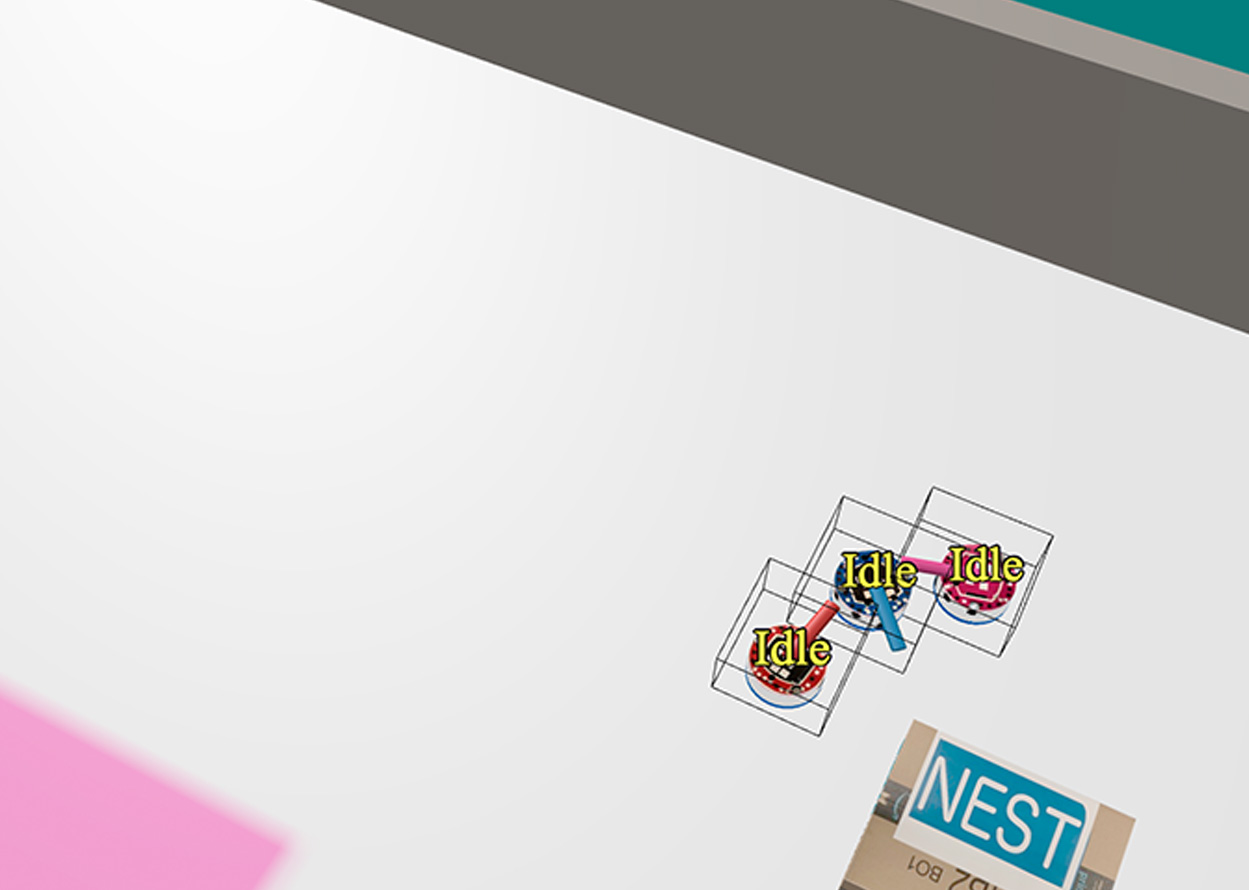}
    \caption{Robot team selection}
    \label{fig-loi:modeS1}
  \end{subfigure}
  \begin{subfigure}{0.49\textwidth}
    \includegraphics[width=\textwidth]{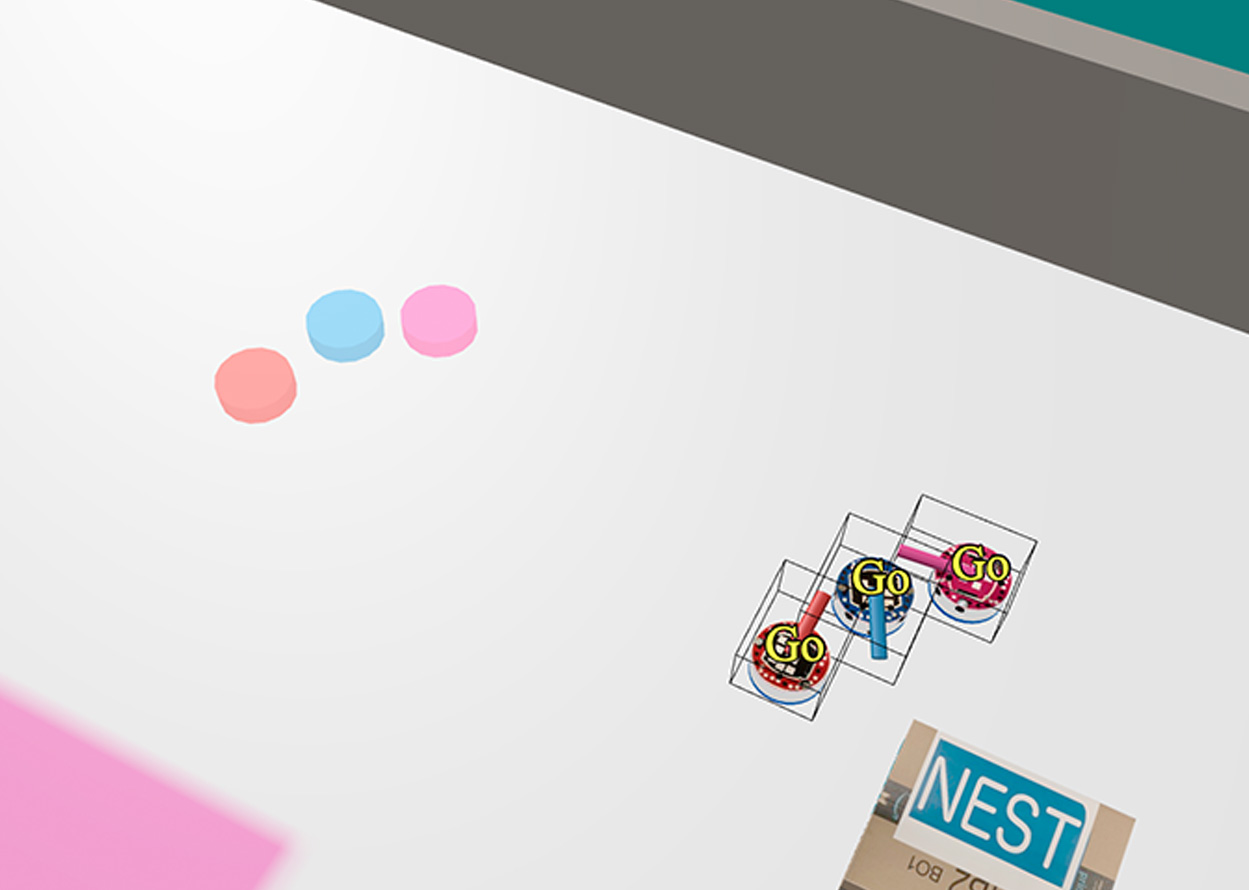}
    \caption{New robot positions}
    \label{fig-loi:modeS2}
  \end{subfigure}
  \begin{subfigure}{0.49\textwidth}
    \includegraphics[width=\textwidth]{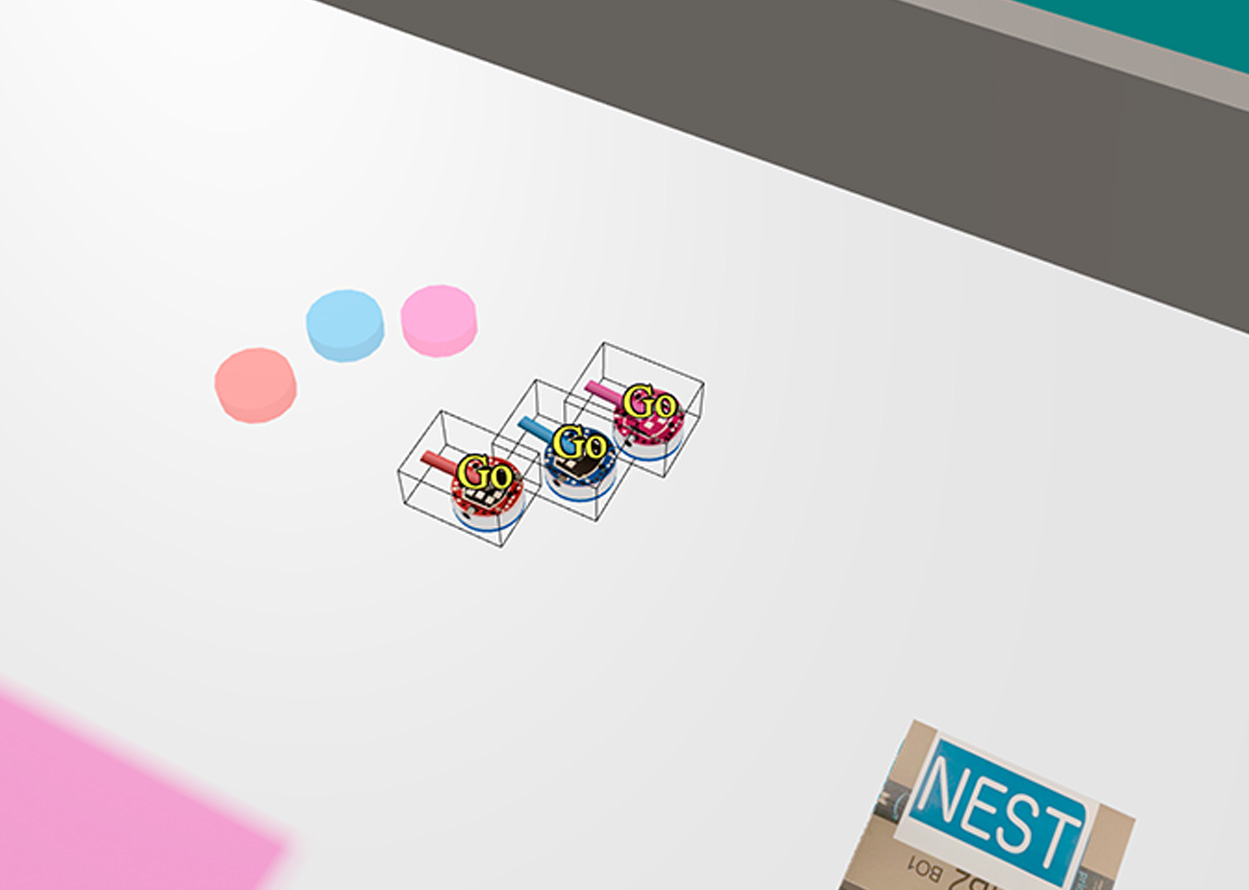}
    \caption{Robots navigating to new position}
    \label{fig-loi:modeS3}
  \end{subfigure}
  \caption{Robot team creation and manipulation by interacting with the interface.}\label{fig-loi:modeS}
\end{figure}

\textbf{Robot Team Selection and Manipulation.} In addition to manipulating a single robot, the operator can select a team of robots by pressing control key and clicking the left mouse button. The goal position is still assigned with a right click. The interface overlays a transparent bounding box over all the selected robots to identify the current selection. If two or more operators have the same robot in their team, then the common robot navigates to the position specified by the last operator without affecting other robots in other teams. Fig.~\ref{fig-loi:modeS1} shows a screenshot in which the selected robots are overlaid with a bounding box. Fig.~\ref{fig-loi:modeS2} shows the goal position visualized as colored virtual objects, one for each of the selected robots. The color of the virtual objects matches the color of the fiducial markers on the body of the robots. Fig.~\ref{fig-loi:modeS3} shows the robots navigating to the goal position.

\subsection{Transparency Modes}
\begin{figure}[t]
    \centering
    \includegraphics[width=\textwidth]{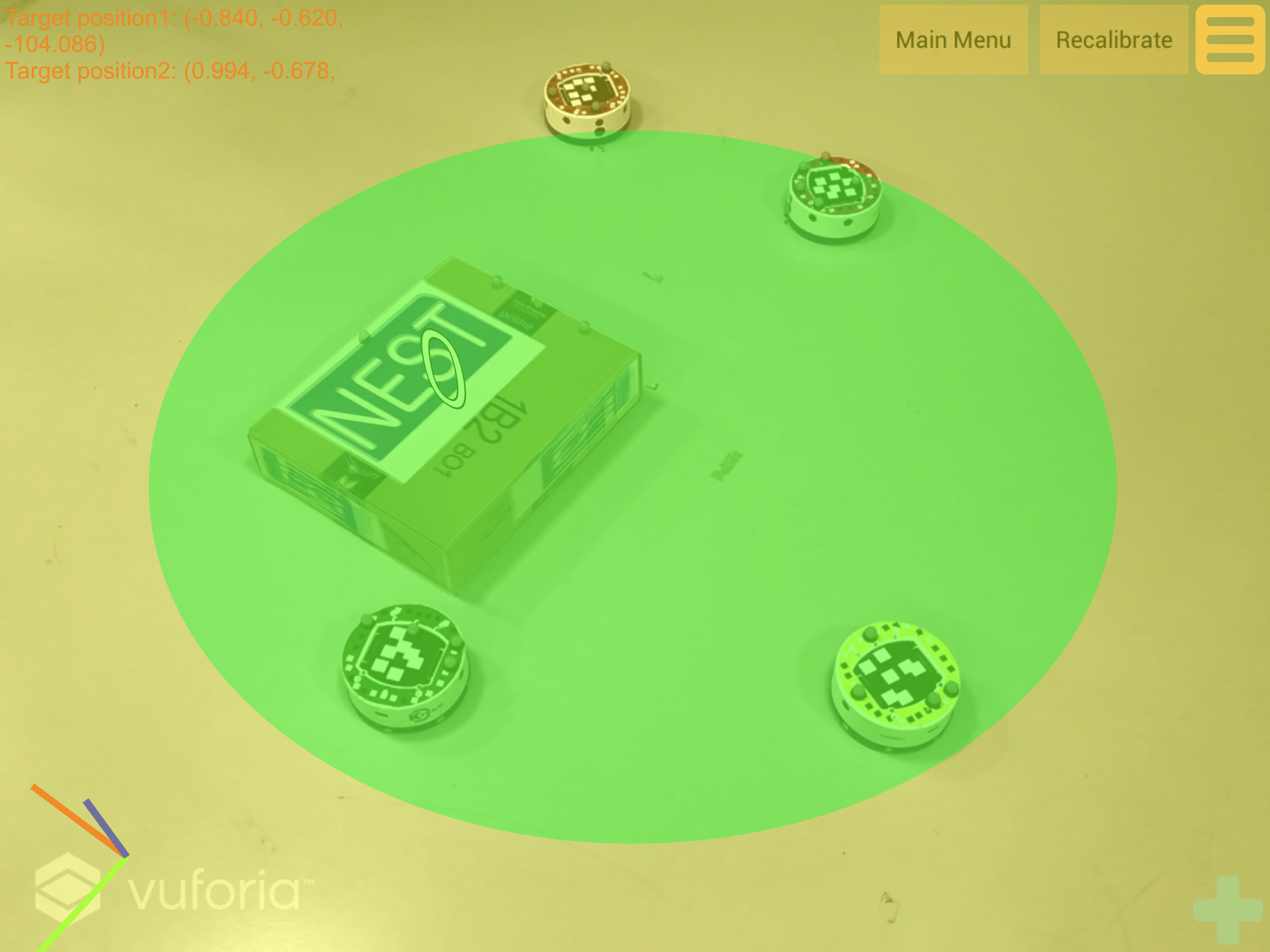}
    \caption{Central and peripheral regions of the field of view. The overlaid green region indicates the central field of view. The overlaid yellow region indicates the peripheral field of view.}
    \label{fig-loi:fov}
\end{figure}

To investigate the role of various elements of the user interface, we endowed our client with the possibility to provide information to the user in several modalities. The main insight in our work is to consider the natural field of view of the human eye (see Fig.~\ref{fig-loi:fov}). We implemented our client to allow for both \emph{central transparency}, i.e., displaying elements in the center of the screen or directly above robots and objects (green region in Fig.~\ref{fig-loi:fov}); and \emph{peripheral transparency}, i.e., relegating interface elements to the borders of the screen (yellow region in Fig.~\ref{fig-loi:fov}). The key difference between central and peripheral transparency is the type and quantity of information displayed. With central transparency, the information is contextual and limited to the robots effectively visible on the screen (which changes as the operator modifies the camera pose). Peripheral transparency, on the other hand, always displays summary information on all the robots and the progress of each task.
\begin{figure}[t]
    \centering
    \includegraphics[width=\textwidth]{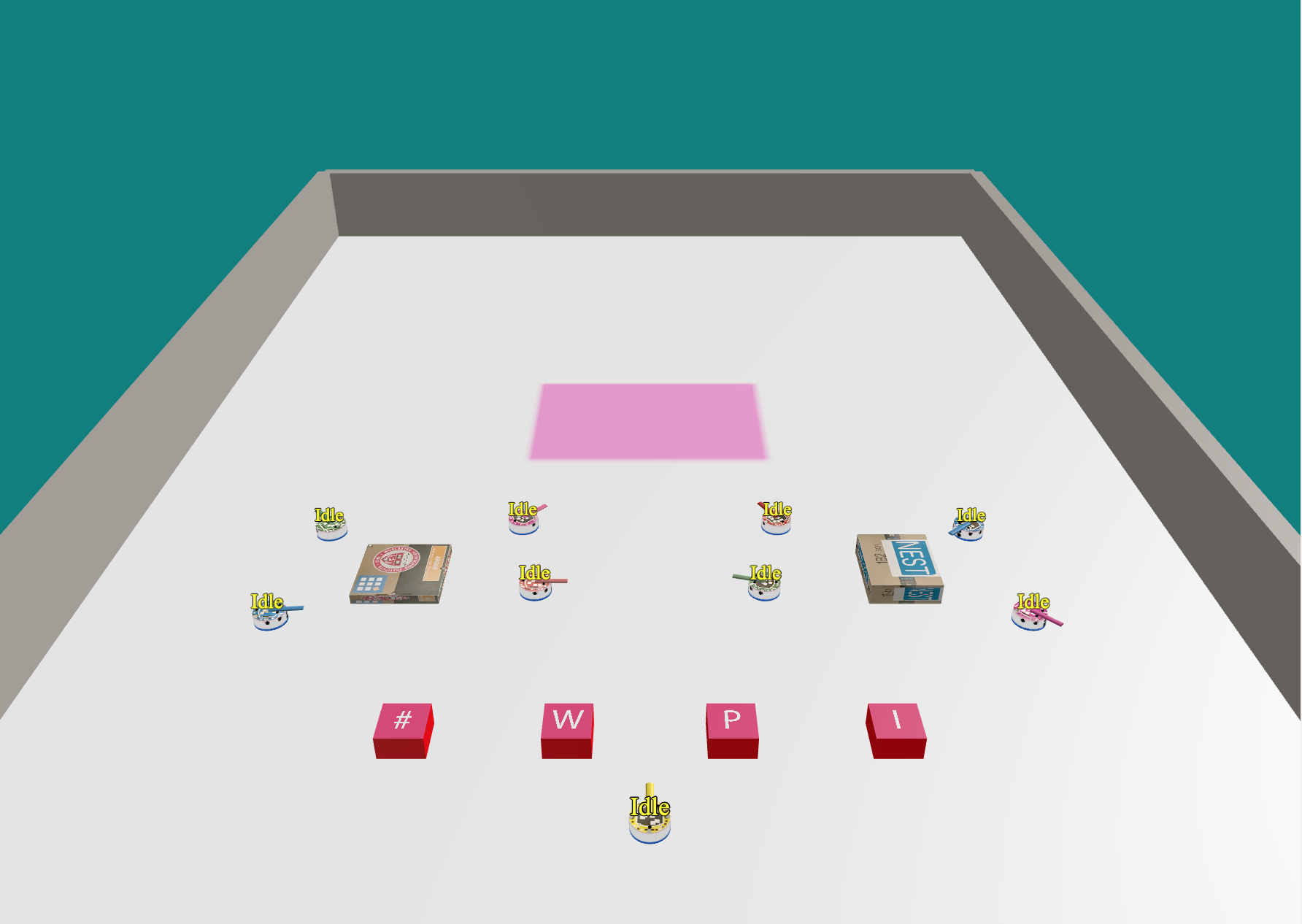}
    \caption{Central transparency showing on-robot status and directional indicator.}
    \label{fig-loi:central}
\end{figure}
\begin{figure}[t]
    \centering
    \includegraphics[width=\textwidth]{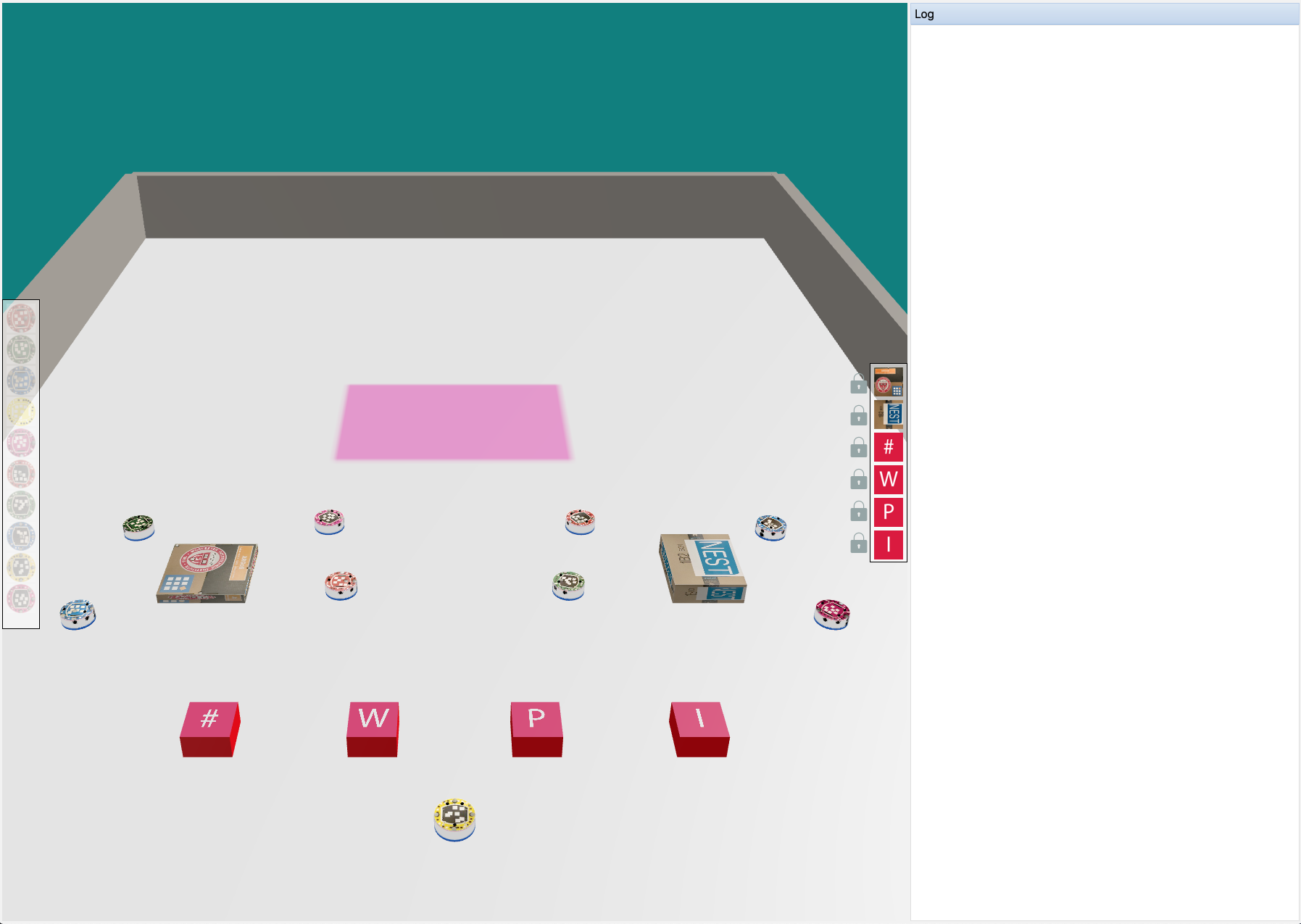}
    \caption{Peripheral transparency mode showing robot panel, object panel and a log (left to right).}
    \label{fig-loi:peripheral}
\end{figure}

The interface can be configured to show or hide every element. For the purposes of our work, we identified four essential “transparency modes”:

\begin{itemize}
\item \textbf{No Transparency (NT).} The interface hides all the information originated by the robots or other operators. The operator can still interact with robots and objects using all the control modalities.
\item \textbf{Central Transparency (CT).} The interface overlays a direction pointer and text to indicate the heading and current task of each robot (as shown in Fig.~\ref{fig-loi:central}). The color of the pointer resembles the color of the fiducial markers on each robot to differentiate between multiple pointers. The robot status displays the current operation executed by the robot corresponding to the states of the collective transport finite state machine (see Fig.~\ref{fig-loi:statemachine}). Additionally, the interface indicates the commands of other operators, to foster shared awareness across operators. This information is available only for entities in the operator's field-of-view. The operator can move around in the environment to view information of other robots and objects that are not in the current field-of-view.
\item \textbf{Peripheral Transparency (PT).} The interface offers a robot panel, an object panel, and a log window containing global information on the system and its constituents (see Fig.~\ref{fig-loi:peripheral}). The robot panel contains one icon for each robot. The panel highlights the icon corresponding to the robots that are moving or performing operator-defined actions. The panel also displays a warning, through a blinking exclamation point, to notify the operators of any fault conditions. These include getting stuck due to an obstacle, and software or hardware failures. The object panel shows all the objects in the environment. The interface highlights the objects currently manipulated by the robots. The panel also provides a functionality to select an object by clicking on the lock icon. An operator can convey their intention of manipulating an object by selecting the lock in the object panel. The interface highlights the lock with a blue icon to signify own selection and a red icon to indicate the selection of another operator. An operator can lock only one object at a time and cannot overwrite the selection of other operators.
\item \textbf{Mixed Transparency (MT).} The interface also allows one to enable both central and peripheral transparency. In this case, the displayed information is a combination of the two transparency modes.
\end{itemize}

\subsection{Communication Modes}
Analogously to transparency modes, the interface also defines different modes for inter-human communication. We classify inter-human communication into direct, indirect, and a combination of both. The communication modes are described as follows.

\begin{itemize}
\item \textbf{No Communication (NC).} In this mode, the operators are completely unable to communicate with each other. The interface hides all the information originating from other operators, such as which robots are being used and which objects are being manipulated.
\item \textbf{Direct Communication (DC).} In this mode, the operators can communicate verbally while performing the task. We established a verbal communication channel using Zoom\footnote{\url{www.zoom.us}}, a video-conferencing application. The operators are allowed to ask for help and strategize at will towards the completion of the task.
\item \textbf{Indirect Communication (IC).} In contrast to direct communication, in this mode the operators cannot verbally communicate their intentions and actions, but they can use the presented transparency modes to communicate indirectly. In this paper, the choice of which transparency mode is active was determined by us at experiment time for the purposes of our study. In a realistic setting, however, each operator is allowed to choose the most appropriate mode.
\item \textbf{Mixed Communication (MC).} In this mode, the operators can communicate both directly and indirectly throughout the duration of the experiment.
\end{itemize}
\section{User Study under Ideal Conditions}
\label{sec-loi:tcstudy}

\subsection{Preliminaries}
The main purpose of this first set of experiments is to validate the usability of the various transparency ($T$) and communication ($C$) modes under ideal conditions in remote interaction ($R$), i.e., with negligible loss of information. We base the experiments on the following main hypotheses.

\noindent \textbf{Hypotheses on the impact of different transparency modes:}
\begin{itemize}
\item \textbf{H$^R_T$1:} Mixed transparency (MT) has the best outcome with respect to other modes.
\item \textbf{H$^R_T$2:} Operators prefer mixed transparency (MT) over other modes.
\item \textbf{H$^R_T$3:} Operators prefer central transparency (CT) over peripheral transparency (PT).
\end{itemize}
\noindent \textbf{Hypotheses on the impact of different communication modes:}
\begin{itemize}
\item \textbf{H$^R_C$1:} Mixed communication (MC) has the best outcome with respect to other modes.
\item \textbf{H$^R_C$2:} Operators prefer mixed communication (MC) over other modes.
\item \textbf{H$^R_C$3:} Operators prefer direct communication (DC) over indirect communication (IC).
\end{itemize}

\begin{figure}[t]
  \centering
  \includegraphics[width=\textwidth]{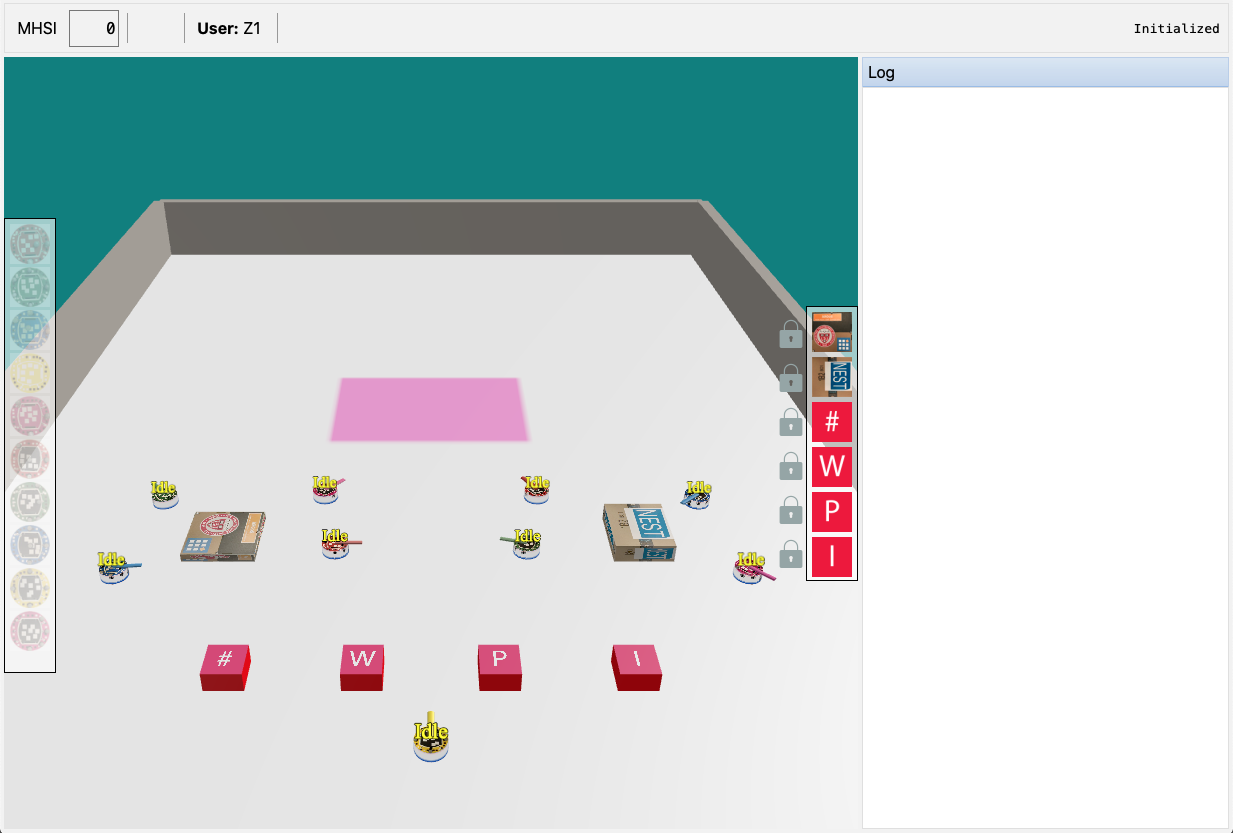}
  \caption{Remote study experiment setup.}
  \label{fig-loi:setup}
\end{figure}

\textbf{Experimental Setup.}
We designed a game scenario (shown in Fig.~\ref{fig-loi:setup}) where the operators were given 9 robots to transport 6 objects (2 big and 4 small) to a goal region. Big objects were worth 2 points each, and small objects were worth 1 point each. The operators had to work as a team to score as many points as possible, over a maximum of 8, in experiments lasting 8 minutes. The operators could move the big objects using the collective transport behavior, or directly use individual robots or team manipulation commands to push the objects.

\textbf{Participant Sample.}
For this user study, we recruited 28 university students. 14 of them (5 female, 9 male), with ages ranging from 19 to 37 years old ($23.28 \pm 4.38$), performed the task four times with a different transparency mode (NT, CT, PT and MT) each time. The other 14 participants (4 female, 10 male), with ages ranging from 18 to 48 years old ($23.64 \pm 7.87$), performed the task four times with a different communication mode (NC, DC, IC and MC) each time. We chosen the teams and the assignments at random. No participant had prior experience with the remote interface.

\textbf{Procedures.}
Each session of the study had two participants and approximately took a total of 105 minutes. After signing the consent form, we explained the task and gave each participant 10 minutes to familiarize with the system. We randomized the order of the tasks and the modalities to reduce the influence of learning effects. After each task, the participants had to answer a subjective questionnaire.

\textbf{Metrics.}
We recorded subjective and objective measures for each participant and each task. We used the following common measures:

\begin{itemize}
\item \textbf{Situational Awareness.} We used the Situational Awareness Rating Technique (SART)~\cite{taylor2017situational} on a 4-point Likert scale~\cite{likert} to assess the awareness of the situation after each task.
\item \textbf{Task Workload.} We used the NASA TLX~\cite{hart1988development} scale on a 4-point Likert scale to compare the perceived workload in each task.
\item \textbf{Trust.} We used the trust questionnaire~\cite{uggirala2004measurement} on a 4-point Likert scale to compare the trust in the interface affected by each transparency mode.
\item \textbf{Quality of Interaction.} We used a custom questionnaire on a 5-point Likert scale to assess the team-level and robot-level interaction. The interaction questionnaire is reported in Fig.~\ref{fig-loi:questionnaire}.
\item \textbf{Performance.} We used the points earned for each task as a metric to scale the performance achieved for each transparency mode.
\item \textbf{Usability.} We asked participants to select the features (log, robot panel, object panel, and on-robot status) they used during the study. Additionally, we asked them to rank the transparency modes from 1 to 4, 1 being the highest rank.
\end{itemize}

\begin{figure}[t]
  \fbox{
    \begin{minipage}{1.0\linewidth}
      \begin{itemize}
      \item[-] Did you understand your \emph{teammate’s intentions}? Were you able to understand why your teammate was taking a certain action?
      \item[-] Could you understand your \emph{teammate’s actions}? Could you understand what your teammate was doing at any particular time?
      \item[-] Could you follow the \emph{progress of the task}? While performing the tasks, were you able to gauge how much of it was pending?
      \item[-] Did you understand what the \emph{robots were doing}? At all times, were you sure how and why the robots were behaving the way they did?
      \item[-] Was the information provided by the interface \emph{clear to understand}?
      \end{itemize}
    \end{minipage}}
  \caption{The subjective questionnaire employed in our user study to assess the quality of interaction of an operator with our interface.}
  \label{fig-loi:questionnaire}
\end{figure}

\begin{table}[h!]
  \centering
  \caption{Results with relationships between transparency modes. The relationships are based on mean ranks obtained through a Friedman Test. The symbol $^*$ denotes significant difference ($p<0.05$) and the symbol $^{**}$ denotes marginally significant difference ($p<0.10$). The symbol $^-$ denotes negative scales where lower ranking is better.}
  \renewcommand{\arraystretch}{1}
  \begin{tabular}{c|c|c|c}
    \hline
    \textbf{Attributes}            & \textbf{Relationship}      & \textbf{$\chi^2(3)$}   & \textbf{$p$-value}    \\ \hline
    \multicolumn{4}{c}{\textbf{SART SUBJECTIVE SCALE}}                                                           \\ \hline\hline
    Instability of Situation$^-$   & NT$>$PT$>$CT$>$MT$^{**}$   & $9.554$                 & $0.023$              \\
    Complexity of Situation$^-$    & NT$>$PT$>$CT$>$MT$^{**}$   & $16.950$                & $0.001$              \\
    Variability of Situation$^-$   & not significant            & $2.452$                 & $0.484$              \\
    Arousal                        & MT$>$CT$>$PT$>$NT$^{**}$   & $8.550$                 & $0.036$              \\
    Concentration of Attention     & MT$>$CT$>$PT$>$NT$^{**}$   & $11.898$                & $0.008$              \\
    Spare Mental Capacity          & not significant            & $2.209$                 & $0.530$              \\
    Information Quantity           & MT$>$CT$>$PT$>$NT$^{**}$   & $12.288$                & $0.006$              \\
    Information Quality            & MT$>$CT$>$PT$>$NT$^{**}$   & $28.758$                & $<0.001$              \\
    Familiarity with Situation     & CT$>$MT$>$PT$>$NT$^{*}$    & $6.276$                 & $0.099$              \\ \hline\hline
    \multicolumn{4}{c}{\textbf{NASA TLX SUBJECTIVE SCALE}}                                                       \\ \hline\hline
    Mental Demand$^-$              & NT$>$PT$>$CT$=$MT$^{**}$   & $10.800$                & $0.013$              \\
    Physical Demand$^-$            & not significant            & $5.634$                 & $0.131$              \\
    Temporal Demand$^-$            & not significant            & $1.760$                 & $0.624$              \\
    Performance                    & not significant            & $6.169$                 & $0.104$              \\
    Effort$^-$                     & PT$>$NT$>$MT$>$CT$^{**}$   & $6.630$                 & $0.085$              \\
    Frustration$^-$                & not significant            & $0.667$                 & $0.881$              \\ \hline\hline
    \multicolumn{4}{c}{\textbf{TRUST SUBJECTIVE SCALE}}                                                          \\ \hline\hline
    Competence                     & MT$>$CT$>$PT$>$NT$^{**}$   & $10.663$                & $0.014$              \\
    Predictability                 & MT$>$CT$>$PT$>$NT$^{**}$   & $19.469$                & $<0.001$             \\
    Reliability                    & MT$>$CT$>$PT$>$NT$^{*}$    & $7.478$                 & $0.058$              \\
    Faith                          & MT$>$CT$>$PT$>$NT$^{**}$   & $15.138$                & $0.002$              \\
    Overall Trust                  & MT$>$CT$>$PT$>$NT$^{**}$   & $18.210$                & $<0.001$             \\
    Accuracy                       & MT$>$CT$>$PT$>$NT$^{**}$   & $10.590$                & $0.014$              \\ \hline\hline
    \multicolumn{4}{c}{\textbf{INTERACTION SUBJECTIVE SCALE}}                                                    \\ \hline\hline
    Teammate's Intent              & MT$>$CT$>$PT$>$NT$^{**}$   & $9.923$                 & $0.019$              \\
    Teammate's Action              & MT$>$CT$>$NT$>$PT$^{**}$   & $8.040$                 & $0.045$              \\
    Task Progress                  & MT$>$CT$>$PT$>$NT$^{*}$    & $6.532$                 & $0.088$              \\
    Robot Status                   & MT$>$CT$>$PT$>$NT$^{**}$   & $15.593$                & $0.001$              \\
    Information Clarity            & CT$>$MT$>$PT$>$NT$^{**}$    & $8.414$                 & $0.038$              \\ \hline\hline
    \multicolumn{4}{c}{\textbf{PERFORMANCE OBJECTIVE SCALE}}                                                     \\ \hline\hline
    Points Scored                  & not significant            & $3.444$                 & $0.328$              \\ \hline
  \end{tabular}
  \label{tab-loi:results-transparency}
  \renewcommand{\arraystretch}{1}
\end{table}

\begin{figure}[t]
  \centering
  \includegraphics[width=0.75\textwidth]{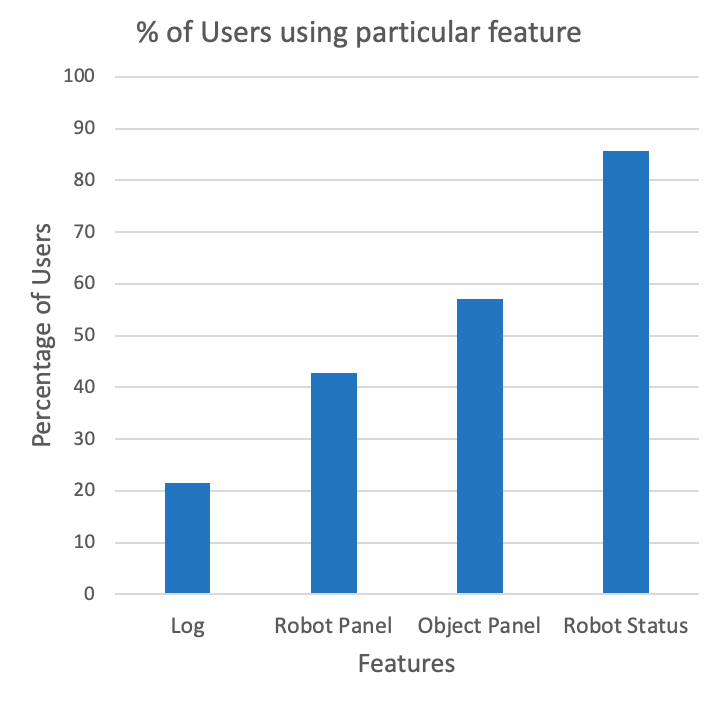}
  \caption{Feature usability in the transparency user study.}
  \label{fig-loi:feature_usability-transparency}
\end{figure}

\begin{figure}[t]
  \centering
  \includegraphics[width=0.75\textwidth]{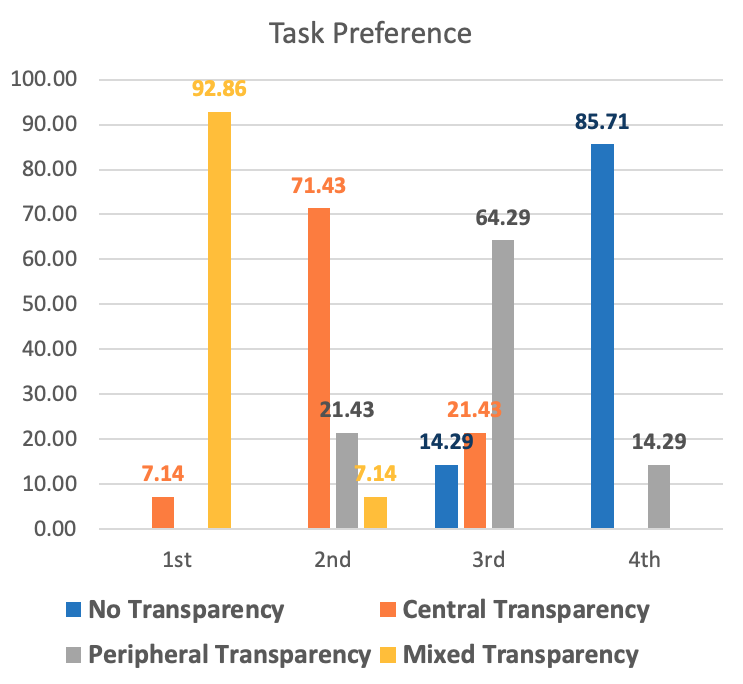}
  \caption{Task preference in the transparency user study.}
  \label{fig-loi:task_usability-transparency}
\end{figure}

\begin{table}[t]
  \centering
  \caption{Ranking scores, in the transparency user study, based on the Borda count. The gray cells indicate the leading scenario for each type of ranking.}
  \renewcommand{\arraystretch}{1}
  \begin{tabular}{c|c|c|c|c}
    \hline\hline
    Borda Count                                                                             & NT & CT    & PT    & MT     \\ \hline\hline
    Based on Collected Data Ranking (Table~\ref{tab-loi:results-transparency})           & 22 & 63.5  & 38    & \greycell{76.5}  \\
    Based on Preference Data Ranking (Fig.~\ref{fig-loi:task_usability-transparency})    & 16 & 40    & 29    & \greycell{55}    \\ \hline
  \end{tabular}
  \label{tab-loi:borda-transparency}
  \renewcommand{\arraystretch}{1}
\end{table}

\begin{table}[h!]
  \centering
  \caption{Results with relationships between communication modes. The relationship are based on mean ranks obtained through Friedman Test. The symbol $^*$ denotes significant difference ($p<0.05$) and the symbol $^{**}$ denotes marginally significant difference ($p<0.10$). The symbol $^-$ denotes negative scales and lower ranking is a good ranking.}
  \renewcommand{\arraystretch}{1}
  \begin{tabular}{c|c|c|c}
    \hline
    \textbf{Attributes}            & \textbf{Relationship}      & \textbf{$\chi^2(3)$}   & \textbf{$p$-value}    \\ \hline
    \multicolumn{4}{c}{\textbf{SART SUBJECTIVE SCALE}}                                                           \\ \hline\hline
    Instability of Situation$^-$   & NC$>$DC$>$IC$>$MC$^{**}$   & $29.105$                & $<0.001$             \\
    Complexity of Situation$^-$    & NC$>$IC$>$DC$>$MC$^{**}$   & $14.921$                & $0.002$              \\
    Variability of Situation$^-$   & NC$>$DC$>$IC$>$MC$^{**}$   & $9.280$                 & $0.026$              \\
    Arousal                        & MC$>$DC$>$IC$>$NC$^{**}$   & $28.240$                & $<0.001$             \\
    Concentration of Attention     & MC$>$DC$>$IC$>$NC$^{**}$   & $24.570$                & $<0.001$             \\
    Spare Mental Capacity          & MC$>$DC$>$IC$>$NC$^{**}$   & $23.579$                & $<0.001$             \\
    Information Quantity           & not significant            & $3.286$                 & $0.350$              \\
    Information Quality            & not significant            & $4.168$                 & $0.244$              \\
    Familiarity with Situation     & MC$>$DC$>$IC$>$NC$^{**}$   & $12.282$                & $0.006$              \\ \hline\hline
    \multicolumn{4}{c}{\textbf{NASA TLX SUBJECTIVE SCALE}}                                                       \\ \hline\hline
    Mental Demand$^-$              & NC$>$IC$>$DC$>$MC$^{**}$   & $21.023$                & $<0.001$             \\
    Physical Demand$^-$            & NC$>$IC$>$DC$>$MC$^{**}$   & $14.870$                & $0.002$              \\
    Temporal Demand$^-$            & NC$>$IC$>$DC$>$MC$^{**}$   & $17.433$                & $0.001$              \\
    Performance                    & MC$>$DC$>$IC$>$NC$^{**}$   & $12.429$                & $0.006$              \\
    Effort$^-$                     & NC$>$IC$>$DC$>$MC$^{**}$   & $25.093$                & $<0.001$             \\
    Frustration$^-$                & NC$>$IC$>$DC$>$MC$^{**}$   & $9.961$                 & $0.019$              \\ \hline\hline
    \multicolumn{4}{c}{\textbf{TRUST SUBJECTIVE SCALE}}                                                          \\ \hline\hline
    Competence                     & MC$>$DC$>$IC$>$NC$^{**}$   & $23.195$                & $<0.001$             \\
    Predictability                 & MC$>$IC$>$DC$>$NC$^{**}$   & $16.059$                & $0.001$             \\
    Reliability                    & MC$>$IC$>$DC$>$NC$^{* }$   & $6.861$                 & $0.076$              \\
    Faith                          & MC$>$DC$>$IC$>$NC$^{**}$   & $13.425$                & $0.004$              \\
    Overall Trust                  & MC$>$DC$>$IC$>$NC$^{**}$   & $17.396$                & $0.001$             \\
    Accuracy                       & MC$>$DC$>$IC$>$NC$^{**}$   & $16.171$                & $0.001$              \\ \hline\hline
    \multicolumn{4}{c}{\textbf{INTERACTION SUBJECTIVE SCALE}}                                                    \\ \hline\hline
    Teammate's Intent              & MC$>$DC$>$IC$>$NC$^{**}$   & $19.848$                & $<0.001$             \\
    Teammate's Action              & MC$>$DC$>$IC$>$NC$^{**}$   & $21.258$                & $<0.001$             \\
    Task Progress                  & MC$>$DC$>$IC$>$NC$^{**}$   & $13.176$                & $0.004$              \\
    Robot Status                   & MC$>$IC$>$DC$>$NC$^{**}$   & $13.991$                & $0.003$              \\
    Information Clarity            & MC$>$IC$>$DC$>$NC$^{**}$    & $25.160$                & $<0.001$             \\ \hline\hline
    \multicolumn{4}{c}{\textbf{PERFORMANCE OBJECTIVE SCALE}}                                                     \\ \hline\hline
    Points Scored                  & not significant            & $3.444$                 & $0.328$              \\ \hline
  \end{tabular}
  \label{tab-loi:results-communication}
  \renewcommand{\arraystretch}{1}
\end{table}

\begin{figure}[t]
  \centering
  \includegraphics[width=0.75\textwidth]{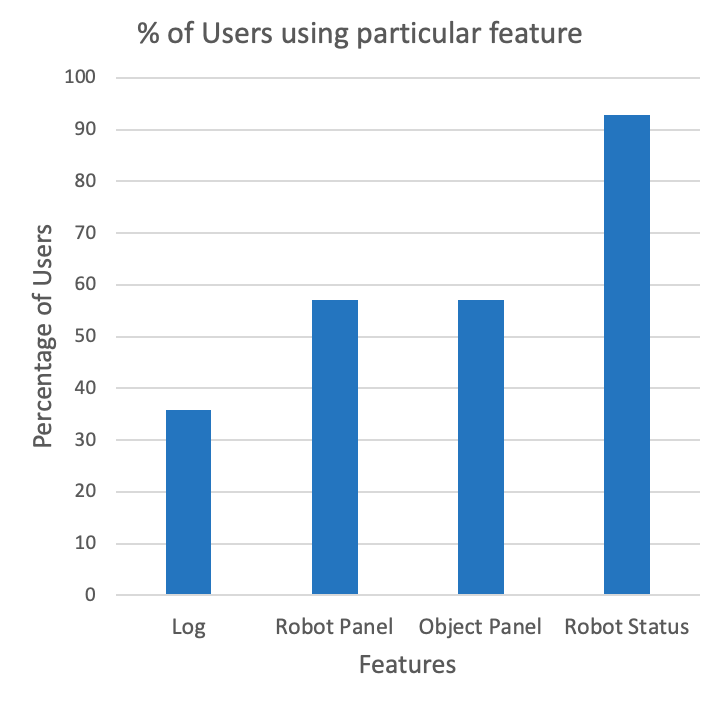}
  \caption{Feature usability in the communication user study.}
  \label{fig-loi:feature_usability-communication}
\end{figure}

\begin{figure}[t]
  \centering
  \includegraphics[width=0.75\textwidth]{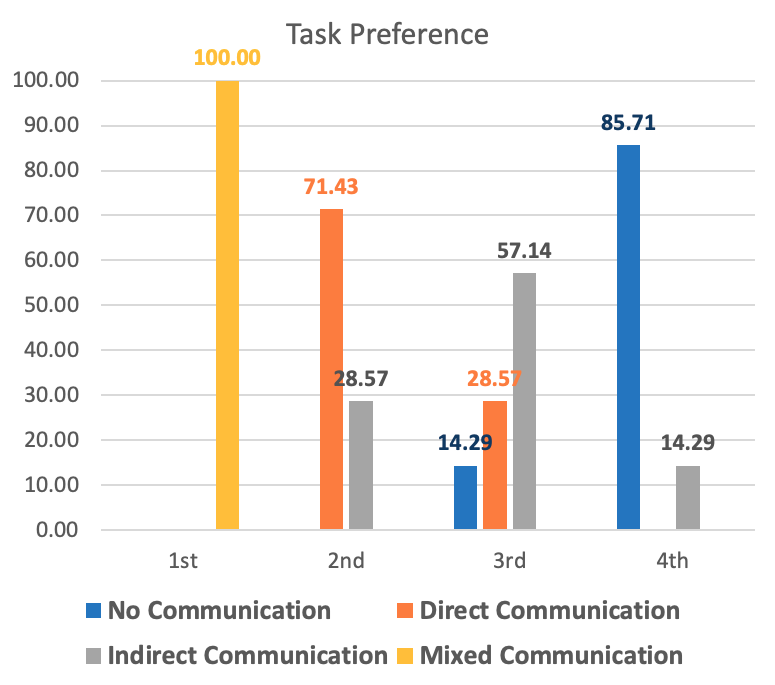}
  \caption{Task preference in the communication user study.}
  \label{fig-loi:task_usability-communication}
\end{figure}

\begin{table}[t]
  \centering
  \caption{Ranking scores, in the communication user study, based on the Borda count. The gray cells indicate the leading scenario for each type of ranking.}
  \renewcommand{\arraystretch}{1}
  \begin{tabular}{c|c|c|c|c}
    \hline\hline
    Borda Count                                                         & NC   & DC    & IC    & MC                            \\ \hline\hline
    Based on Collected Data Ranking (Table~\ref{tab-loi:results-communication})           & 24 & 67  & 53    & \greycell{96}  \\
    Based on Preference Data Ranking (Fig.~\ref{fig-loi:task_usability-communication})    & 16 & 38  & 30    & \greycell{56}    \\ \hline
  \end{tabular}
  \label{tab-loi:borda-communication}
  \renewcommand{\arraystretch}{1}
\end{table}

\subsection{Analysis and Discussion}
\label{sec:discussion}

\subsubsection{Collected Data}
\textbf{Transparency Data.} Table~\ref{tab-loi:results-transparency} shows the summarized results for all the subjective scales and the objective performance. We used the Friedman test~\cite{friedman1937use} to analyze the data and assess the significance between different modes of transparency. We derived a ranking based on the mean ranks for all the attributes that showed statistical significance ($p<0.05$) or marginal significance ($p<0.10$). Fig.~\ref{fig-loi:feature_usability-transparency} shows the percentage of operators using a particular feature. Fig.~\ref{fig-loi:task_usability-transparency} shows the percentage of people ranking a task based on their choice. We used the Borda count~\cite{black1976partial} method for calculating the overall ranking of the collected data and transparency mode usability data. We inverted the ranking of the negative scales for calculating the Borda count scores. Table~\ref{tab-loi:borda-transparency} shows the results of the Borda count for each category.

\textbf{Communication Data.} Table~\ref{tab-loi:results-communication} shows the summarized results of the communication user study. We analyzed the data using the Friedman test~\cite{friedman1937use} to assess the significant relationships among different modes of communication. We used statistical significance ($p<0.05$) and marginal significance ($p<0.10$) to derive a ranking based on their mean ranks. Fig.~\ref{fig-loi:feature_usability-communication} shows the percentage of operators using a particular feature. Fig.~\ref{fig-loi:task_usability-communication} shows the percentage of people ranking task based on their choice. Using the Borda count method, we derived an overall ranking based on the collected data and the user preference data (shown in Table~\ref{tab-loi:borda-communication}).We inverted the ranking of the negative scales for the Borda count scores.

\subsubsection{Transparency Modes}
Table~\ref{tab-loi:borda-transparency} shows that mixed transparency (MT) is the best transparency mode in terms of usability, supporting hypotheses \textbf{H$^R_T$1} and \textbf{H$^R_T$2}. From the results, central transparency (CT) dominates peripheral transparency (PT), supporting hypothesis \textbf{H$^R_T$3}. In addition to this, we also analyzed the modes of transparency based on the sub-scales of the subjective data and further analysed for each mode as follows.  

\textbf{Mixed Transparency.} This mode is the overall best choice for the operators. The results suggest that this mode provides the operators with the best situational awareness, measured in terms of least instability of situation, complexity of situation, best information arousal, level of concentration, information quality, and information quantity. Through this transparency mode, the operators had the most information about actions and intentions of teammates and robots, as well as of the task progress. This led the operators to report the highest trust across all trust sub-scales.

\textbf{Central Transparency.} This mode is the second best choice after mixed transparency. The operators had the best familiarity and clarity in terms of information provided by the interface. The operators experienced the lowest mental load and reported the least effort in performing the task. Fig.~\ref{fig-loi:feature_usability-transparency} supports these findings as 92\% (13 out of 14 operators) indicated the on-robot status as the most useful feature.

\textbf{Peripheral Transparency.} The operators reported peripheral transparency as the most cumbersome mode. The operators experienced the lowest awareness, which caused degraded trust. The operators reported that the mode was merely better than no transparency (NT), because the presence of \emph{some} information is still better than \emph{no} information.

\textbf{Comparison with Proximal Interaction.} Overall, the conclusions of this study are in line those we reported for proximal interaction. However, the results in this paper are more substantial compared to what we observed for proximal interaction. Unlike proximal interaction, mixed transparency in remote interaction was the clear winner, both from the collected data ranking and the preference data ranking (see Table.~\ref{tab-loi:borda-transparency}). Central transparency not only outperformed peripheral transparency in remote interaction, but dominated the results when compared to the findings of the study with proximal interaction. We speculate that this difference is due to the fact that, in proximal interaction, the operators had to devote effort to avoid bumping into robots and other operators while walking. This made the operators alert and anxious, affecting their focus on the information offered by the interface and the transparency modes. In remote interaction, as there was no need to physically move, the operators could focus on the displayed information more effectively.

\begin{figure}[t]
  \centering
  \includegraphics[width=0.75\textwidth]{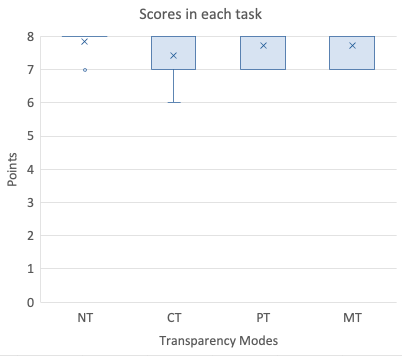}
  \caption{Task performance for each transparency mode.}
  \label{fig-loi:performance-transparency}
\end{figure}

\begin{figure}[h!]
  \centering
  \includegraphics[width=0.75\textwidth]{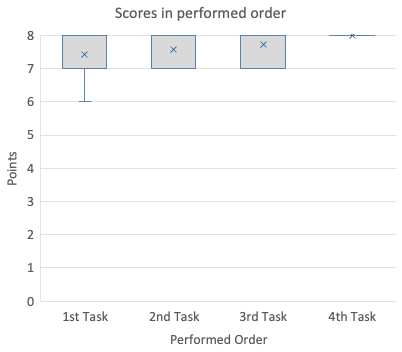}
  \caption{Learning effect in the transparency user study based on points scored.}
  \label{fig-loi:learning-transparency}
\end{figure}

\begin{figure}[t]
  \centering
  \includegraphics[width=0.75\textwidth]{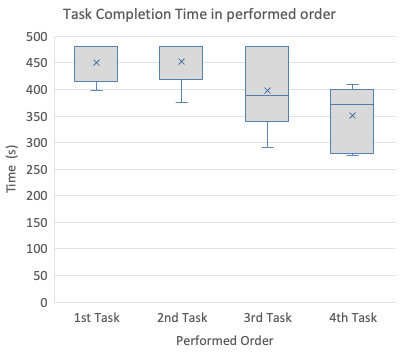}
  \caption{Learning effect in the transparency user study based on time taken to complete the task.}
  \label{fig-loi:learningtime-transparency}
\end{figure}

Our experiments did not reveal a substantial difference in performance across transparency modes. We hypothesize that this lack of difference is due to the learning effect across the four runs that each team had to perform. Fig.~\ref{fig-loi:performance-transparency} shows the performance in each task and Fig.~\ref{fig-loi:learning-transparency} reports the increase in performance due to the task order (learning effect). As most of the teams were able to complete the task in less than 8 minutes, Fig.~\ref{fig-loi:learningtime-transparency} shows the decrease in time taken to complete the task in order of the performed task, indicating the impact of the learning effect. 

\subsubsection{Modes of Communication}
Table~\ref{tab-loi:borda-communication} suggests that mixed communication (MC) is the best mode of communication, both in terms of usability preference and in terms of the data collected during the user study, supporting hypotheses \textbf{H$^R_C$1} and \textbf{H$^R_C$2}. In addition, direct communication (DC) outperformed indirect communication (IC), confirming hypothesis \textbf{H$^R_C$3}. We also analysed the modes of communication based on the sub-scales of the subjective data and further analysed for each mode.

\textbf{Mixed Communication.} Mixed communication was recognized as the best mode, not only based on the Borda count but also looking at the results of the subjective data. This mode had the best situational awareness, trust in the system, and interaction with the robots and the operator, while having the lowest task load.

\textbf{Direct Communication.} This mode was the second best. It outperformed indirect communication in terms of information awareness and communication with the other operator (operator-level information), resulting in better trust in the system and lower workload with respect to indirect communication.

\textbf{Indirect Communication.} This mode was the third best choice. This mode proved to be better in conveying robot-level information, thus allowing the operator to better understand and predict robot actions, when compared to direct communication. This made the operators trust this mode more in terms of predictability and reliability, but at the cost of experiencing higher workload in comparison to mixed communication and direct communication.

\textbf{Comparison with Proximal Interaction.} Analogously to what we said about transparency, these observations are in line with the results of the proximal interaction study \cite{Patel2021}. However, the results of this study were more decisive with respect to the proximal interaction study. Also in this case, we observed that the proximal interaction made the operators alert and anxious about robots and the other operator. Also, as the operators had to physically walk around other robots, the interaction felt at times cumbersome. This observation is supported by workload results of the proximal interaction studies in our previous work, indicating high workload experienced in all modes of communication. In contrast, the results of workload in remote interaction showed significant difference between communication modes.

\begin{figure}[t]
  \centering
  \includegraphics[width=0.75\textwidth]{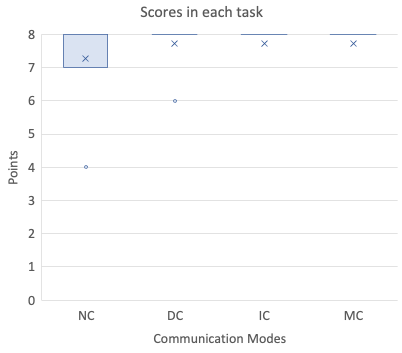}
  \caption{Task performance for each communication mode.}
  \label{fig-loi:performance-communication}
\end{figure}

\begin{figure}[h!]
  \centering
  \includegraphics[width=0.75\textwidth]{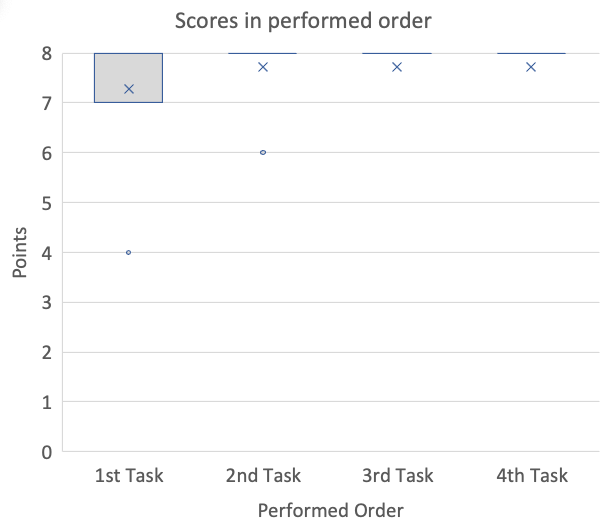}
  \caption{Learning effect in the communication user study.}
  \label{fig-loi:learning-communication}
\end{figure}

\begin{figure}[t]
  \centering
  \includegraphics[width=0.75\textwidth]{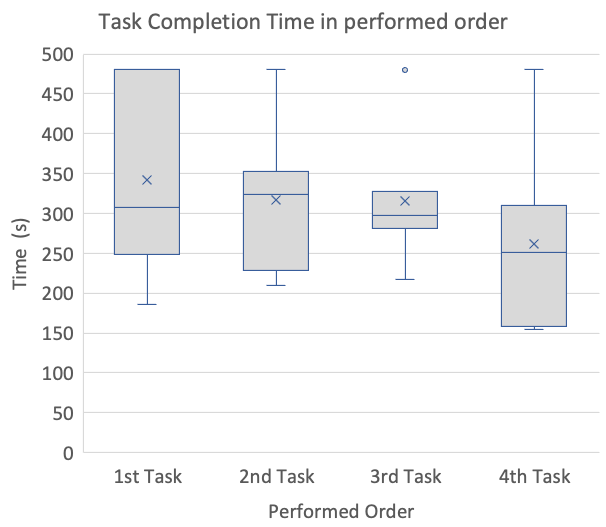}
  \caption{Learning effect in the communication user study.}
  \label{fig-loi:learningtime-communication}
\end{figure}

Our experiments did not reveal a significant difference in performance across communication modes. Similarly to what we discussed for transparency, we hypothesize that this lack of difference is due to the learning effect across the four runs that each team had to perform. Fig.~\ref{fig-loi:performance-communication} indicates the points earned by the operators in each task and Fig.~\ref{fig-loi:learning-communication} shows the learning effect as the increase in points earned in order of the performed task. As most of the operator teams were able to complete the task earlier than 8 minutes, Fig.~\ref{fig-loi:learningtime-communication} shows the decrease in time taken to complete the task in order of the performed task as a clear indicator of the learning effect.

\section{User Study with Information Loss}
\label{sec-loi:informationloss}
The study presented so far was based on the assumption that the information flow was fast and continuous for every operator. This was possible because all the users involved in our experimental evaluation had fast, stable Internet connections that showed no issues. However, in remote operations, fast and stable connectivity cannot be taken for granted.

For this reason, we investigate the role that intermittent information flow plays in the efficiency of remote multi-human multi-robot interaction. In this paper, we measure information loss as the time elapsed between two updates of the graphical user interface. In other words, we define information loss as the inverse of the frame rate. With operators and robots in separate environments, it is likely for the operators to experience different levels of information loss. When this happens, we speak of \emph{heterogeneous} information loss.

For the purposes of our study, we categorize information loss in two ranges of usability. The \emph{high usability range (U$_H$)} corresponds to levels of information loss that cause negligible discomfort in the operators that experience it. Conversely, we are in \emph{low usability range (U$_L$)} when the level of information loss is such that an operator cannot ignore its presence, experiencing some sort of discomfort.

In general, the exact extent of these ranges changes with the operators. We thus split our study in two parts. In the \emph{pilot study} (Sec.~\ref{sec:pilotstudy}), we investigate the extent of the usability ranges in experiments that involve a single operator. Next, in the \emph{main study} (see Sec.~\ref{sec:mainstudy}), we turn to multiple operators and assess the effect of heterogeneous information loss, using the homogeneous case as a baseline reference.


\subsection{Information Loss Pilot Study}
\label{sec:pilotstudy}

\textbf{Experimental Setup.}
For our pilot study with a single operator we used the game scenario presented in Sec.~\ref{sec-loi:tcstudy} (see Fig.~\ref{fig-loi:setup}). The operator was tasked with performing half of the game: moving 1 big object and 2 small objects. In contrast to the previous game, we set no time limit to complete the task, instead declaring completion when the required objects reached the goal region. Every participant had to perform the task 6 times with different levels of information loss each time. The levels spanned from \unit[0]{s} to \unit[2.5]{s} in increments of \unit[0.5]{s}. To compensate for possible learning effects or other confusing factors, we determined different level orderings:
\begin{itemize}
\item \textbf{Increasing order}: information loss increases with every task.
\item \textbf{Decreasing order}: the information loss decreases with every task.
\item \textbf{Random 1}: information loss is in the order $\{0, 2.5, 0.5, 2, 1, 1.5\}$ s.
\item \textbf{Random 2}: the reverse order with respect to Random 1.
\end{itemize}

\textbf{Participant Sample.}
We recruited 20 university students (7 females, 13 males) with ages ranging from 18 to 31 years old ($22.75 \pm 3.57$). All participants were randomly assigned one task ordering. Each participant performed the 6 tasks in the determined order. No participant had prior experience with the remote interface.

\textbf{Pilot Study Procedure.}
Each session of the study took approximately 90 minutes. After signing the consent form, we explained the task setup and gave the participant 12 minutes to familiarize with the system. After each task, the participant had to answer a subjective questionnaire.

\textbf{Metrics.}
We recorded the subjective and objective measures for each participant for each task. The performance of the operator was measured as time taken to complete a task. We used the NASA TLX~\cite{hart1988development} scale on a 10-point Likert scale to compare the perceived workload in each task. In addition to the workload questionnaire, the participants were requested to report the experienced discomfort on a 10-point Likert Scale, followed by a comment box for free-form description of the type of discomfort experienced.

\begin{figure}[t!]
  \centering
  \includegraphics[width=0.75\textwidth]{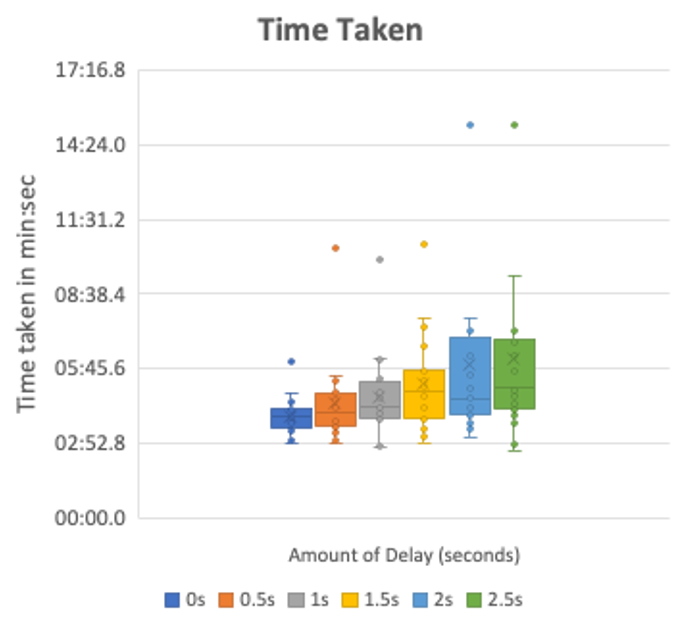}
  \caption{Box plot for performance, in time taken to complete the task. Lower is better.}
  \label{fig-loi:performance}
\end{figure}

\begin{figure}[h!]
  \centering
  \includegraphics[width=0.75\textwidth]{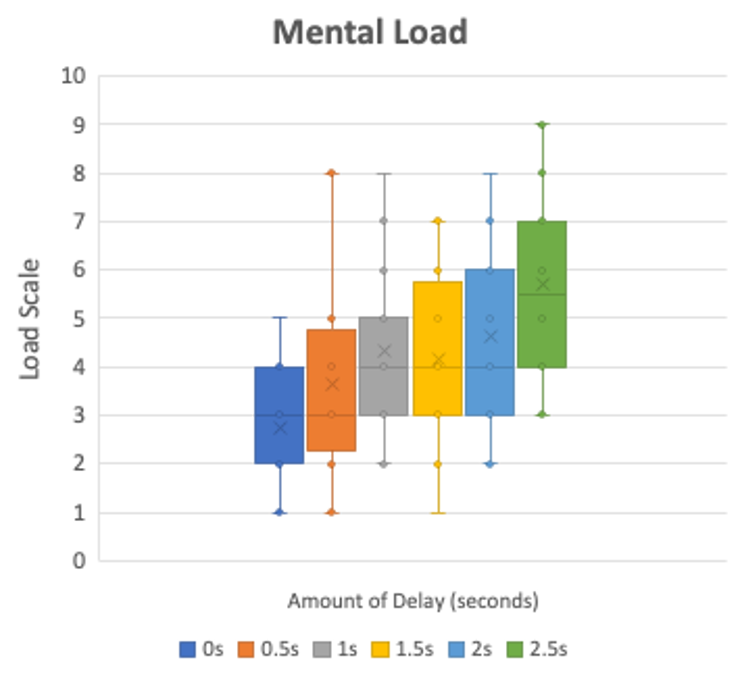}
  \caption{Box plot for reported mental load. Lower is better.}
  \label{fig-loi:mental}
\end{figure}

\begin{figure}[h!]
  \centering
  \includegraphics[width=0.75\textwidth]{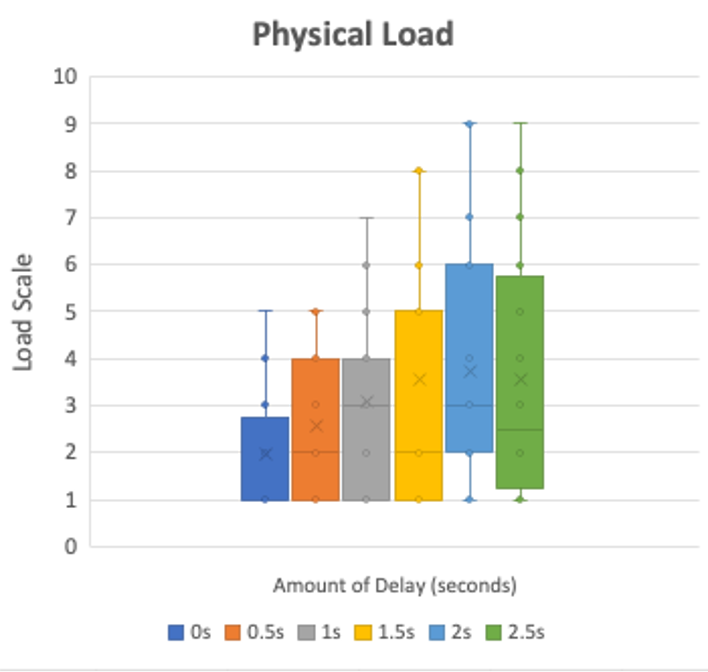}
  \caption{Box plot for reported physical load. Lower is better.}
  \label{fig-loi:physical}
\end{figure}

\begin{figure}[h!]
  \centering
  \includegraphics[width=0.75\textwidth]{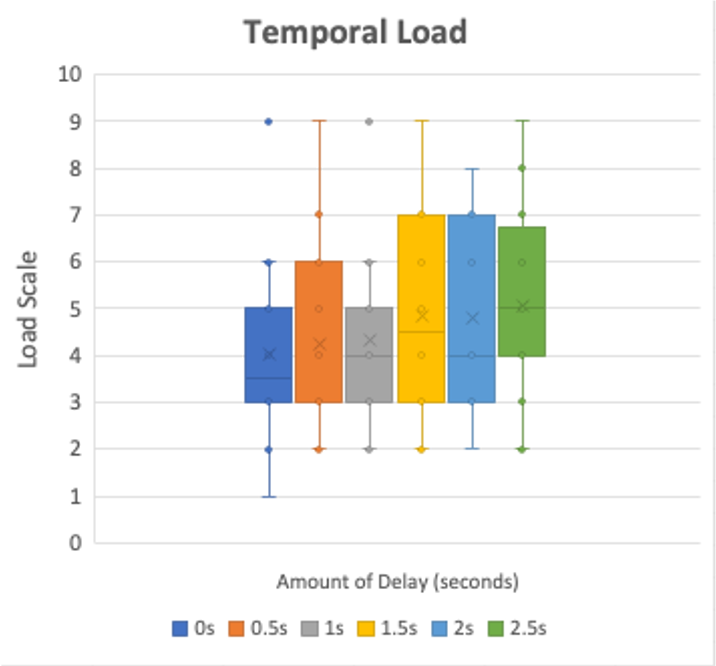}
  \caption{Box plot for reported temporal load. Lower is better.}
  \label{fig-loi:temporal}
\end{figure}

\begin{figure}[h!]
  \centering
  \includegraphics[width=0.75\textwidth]{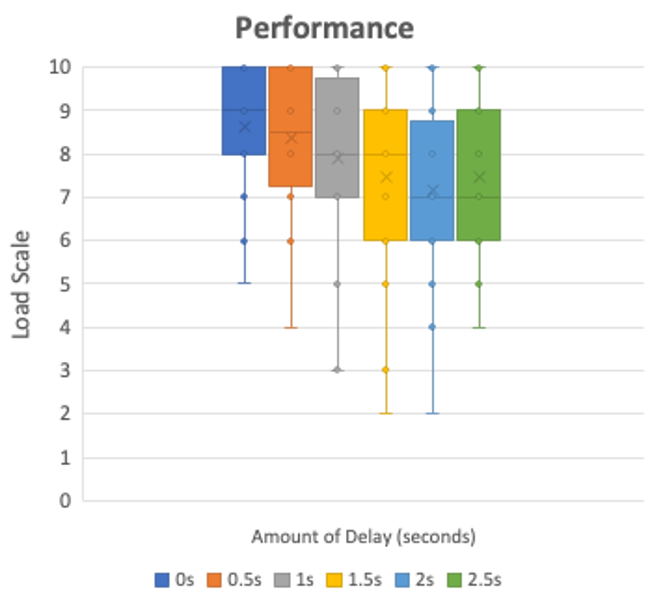}
  \caption{Box plot for reported perceived performance. Higher is better.}
  \label{fig-loi:pp}
\end{figure}

\begin{figure}[h!]
  \centering
  \includegraphics[width=0.75\textwidth]{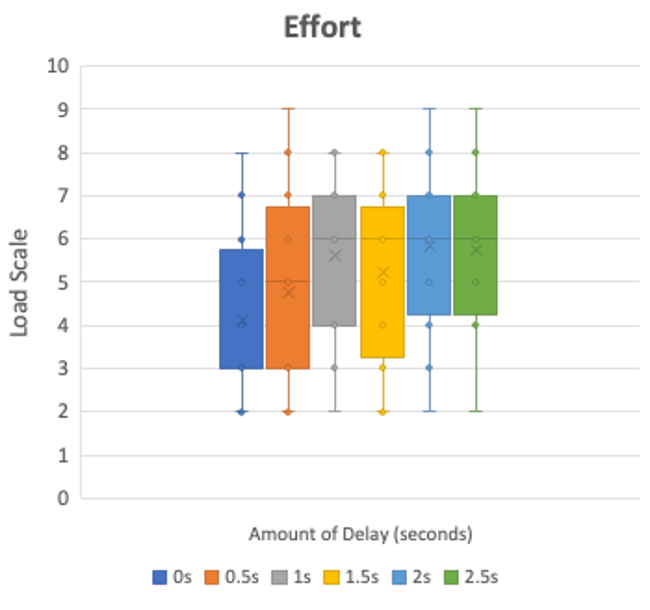}
  \caption{Box plot for reported effort. Lower is better.}
  \label{fig-loi:effort}
\end{figure}

\begin{figure}[h!]
  \centering
  \includegraphics[width=0.75\textwidth]{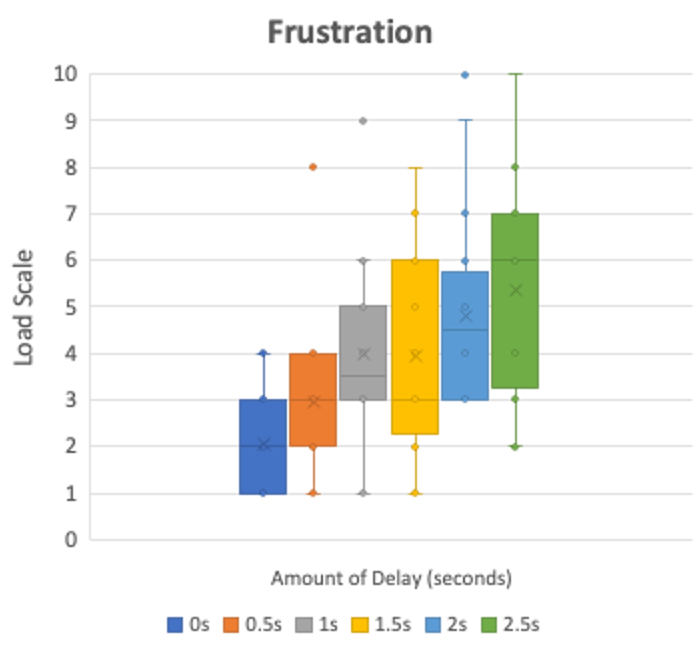}
  \caption{Box plot for reported frustration. Lower is better.}
  \label{fig-loi:frustration}
\end{figure}

\begin{figure}[h!]
  \centering
  \includegraphics[width=0.75\textwidth]{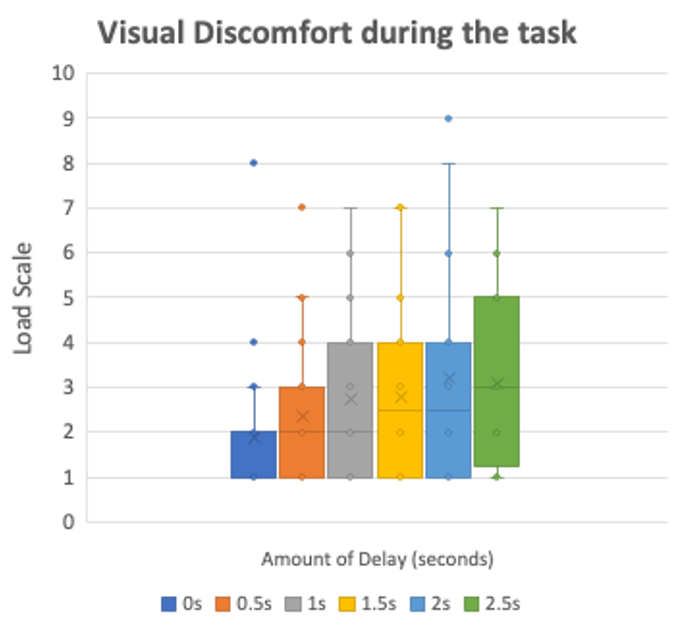}
  \caption{Box plot for reported visual discomfort. Lower is better.}
  \label{fig-loi:vd}
\end{figure}

\textbf{Results.}
For each item in the NASA TLX scale, we report a significance matrix based on the Friedman test to identify the two ranges of usability. The results are shown in Tables~\ref{tab-loi:sm-performance}-\ref{tab-loi:sm-visual}. The green cells in these tables indicate the high usability range and the red cells indicate the low usability range. We also superimposed the usability ranges in Table~\ref{tab-loi:sm-overlaid}. From our data, we estimate the high usability range between \unit[0]{s} and \unit[0.5]{s}, and the low usability range between \unit[2]{s} and \unit[2.5]{s}. For the upcoming main study on information loss (Sec.~\ref{sec:mainstudy}), we took the midpoints of these ranges (\unit[0.25]{s} and \unit[2.25]{s}). Figures~\ref{fig-loi:performance}-\ref{fig-loi:vd} report the box plots of the recorded readings for the respective metrics.

\begin{table}[t]
  \centering
  \caption{Significance matrix for differences in performance between levels of information loss. The shaded regions indicate the two ranges of usability. The cell entries are the $p$-values based on the Friedman test. The empty cells represent a comparison with no significant difference.}
  \renewcommand{\arraystretch}{1.1}
  \begin{tabular}{c||C{1.2cm}|C{1.2cm}|C{1.2cm}|C{1.2cm}|C{1.2cm}|C{1.2cm}}
    \hline\hline
    \textbf{Performance}     &   \textbf{0s}    &   \textbf{0.5s}   &   \textbf{1s}     &   \textbf{1.5s}   &   \textbf{2s}     &   \textbf{2.5s}\\ \hline\hline
    \textbf{0s}              &   \blackcell     &   \greencell      &   0.007           &   0.007           &   $<$0.001          &   0.002        \\ \hline
    \textbf{0.5s}            &   \greencell     &   \blackcell      &                   &                   &   0.025           &    0.025       \\ \hline
    \textbf{1s}              &   0.007          &                   &   \blackcell      &                   &   0.025           &   0.025        \\ \hline
    \textbf{1.5s}            &   0.007          &                   &                   &   \blackcell      &   \redcell        &   \redcell     \\ \hline
    \textbf{2s}              &   $<$0.001         &   0.025           &   0.025           &   \redcell        &   \blackcell      &   \redcell     \\ \hline
    \textbf{2.5s}            &   0.002          &   0.025           &   0.025           &   \redcell        &   \redcell        &   \blackcell   \\ \hline
  \end{tabular}
  \label{tab-loi:sm-performance}
  \renewcommand{\arraystretch}{1}
\end{table}

\begin{table}[t!]
  \centering
  \caption{Significance matrix for differences in mental load between levels of information loss. The shaded regions indicate the two ranges of usability. The cell entries are the $p$-values based on the Friedman test. The empty cells represent a comparison with no significant difference.}
  \renewcommand{\arraystretch}{1.1}
  \begin{tabular}{c||C{1.2cm}|C{1.2cm}|C{1.2cm}|C{1.2cm}|C{1.2cm}|C{1.2cm}}
    \hline\hline
    \textbf{ML}              &   \textbf{0s}    &   \textbf{0.5s}   &   \textbf{1s}     &   \textbf{1.5s}   &   \textbf{2s}     &   \textbf{2.5s}\\ \hline\hline
    \textbf{0s}              &   \blackcell     &   \greencell      &   $<$0.001          &   0.008           &   $<$0.001          &   $<$0.001       \\ \hline
    \textbf{0.5s}            &   \greencell     &   \blackcell      &   0.008           &                   &   0.012           &    0.003       \\ \hline
    \textbf{1s}              &   $<$0.001         &   0.008           &   \blackcell      &                   &                   &    0.018       \\ \hline
    \textbf{1.5s}            &   0.008          &                   &                   &   \blackcell      &   0.033           &    0.005       \\ \hline
    \textbf{2s}              &   $<$0.001         &   0.012           &                   &   0.033           &   \blackcell      &    0.002       \\ \hline
    \textbf{2.5s}            &   $<$0.001         &   0.003           &   0.018           &   0.005           &   0.002           &   \redcell     \\ \hline
  \end{tabular}
  \label{tab-loi:sm-mental}
  \renewcommand{\arraystretch}{1}
\end{table}

\begin{table}[t!]
  \centering
  \caption{Significance matrix for differences in physical load between levels of information loss. The shaded regions indicate the two ranges of usability. The cell entries are the $p$-values based on the Friedman test. The empty cells represent a comparison with no significant difference.}
  \renewcommand{\arraystretch}{1.1}
  \begin{tabular}{c||C{1.2cm}|C{1.2cm}|C{1.2cm}|C{1.2cm}|C{1.2cm}|C{1.2cm}}
    \hline\hline
    \textbf{PL}              &   \textbf{0s}    &   \textbf{0.5s}   &   \textbf{1s}     &   \textbf{1.5s}   &   \textbf{2s}     &   \textbf{2.5s}\\ \hline\hline
    \textbf{0s}              &   \greencell     &   0.007           &   0.004           &   0.002           &   $<$0.001          &   $<$0.001       \\ \hline
    \textbf{0.5s}            &   0.007          &   \blackcell      &   0.008           &   0.004           &   0.02            &    0.033       \\ \hline
    \textbf{1s}              &   0.004          &   0.008           &   \blackcell      &   \redcell        &   \redcell        &    \redcell    \\ \hline
    \textbf{1.5s}            &   0.002          &   0.004           &   \redcell        &   \blackcell      &   \redcell        &    \redcell    \\ \hline
    \textbf{2s}              &   $<$0.001         &   0.02            &   \redcell        &   \redcell        &   \blackcell      &    \redcell    \\ \hline
    \textbf{2.5s}            &   $<$0.001         &   0.033           &   \redcell        &   \redcell        &   \redcell        &   \blackcell   \\ \hline
  \end{tabular}
  \label{tab-loi:sm-physical}
  \renewcommand{\arraystretch}{1}
\end{table}

\begin{table}[t!]
  \centering
  \caption{Significance matrix for differences in temporal load between levels of information loss. The shaded regions indicate the two ranges of usability. The cell entries are the $p$-values based on the Friedman test. The empty cells represent a comparison with no significant difference.}
  \renewcommand{\arraystretch}{1.1}
  \begin{tabular}{c||C{1.2cm}|C{1.2cm}|C{1.2cm}|C{1.2cm}|C{1.2cm}|C{1.2cm}}
    \hline\hline
    \textbf{TL}              &   \textbf{0s}    &   \textbf{0.5s}   &   \textbf{1s}     &   \textbf{1.5s}   &   \textbf{2s}     &   \textbf{2.5s}\\ \hline\hline
    \textbf{0s}              &   \blackcell     &   \greencell      &   \greencell      &   0.013           &                   &                \\ \hline
    \textbf{0.5s}            &   \greencell     &   \blackcell      &   \greencell      &   0.046           &                   &                \\ \hline
    \textbf{1s}              &   \greencell     &   \greencell      &   \blackcell      &   \redcell        &   \redcell        &    \redcell    \\ \hline
    \textbf{1.5s}            &   0.013          &   0.046           &   \redcell        &   \blackcell      &   \redcell        &    \redcell    \\ \hline
    \textbf{2s}              &                  &                   &   \redcell        &   \redcell        &   \blackcell      &    \redcell    \\ \hline
    \textbf{2.5s}            &                  &                   &   \redcell        &   \redcell        &   \redcell        &   \blackcell   \\ \hline
  \end{tabular}
  \label{tab-loi:sm-temporal}
  \renewcommand{\arraystretch}{1}
\end{table}

\begin{table}[t!]
  \centering
  \caption{Significance matrix for differences in perceived performance between levels of information loss. The shaded regions indicate the two ranges of usability. The cell entries are the $p$-values based on the Friedman test. The empty cells represent a comparison with no significant difference.}
  \renewcommand{\arraystretch}{1.1}
  \begin{tabular}{c||C{1.2cm}|C{1.2cm}|C{1.2cm}|C{1.2cm}|C{1.2cm}|C{1.2cm}}
    \hline\hline
    \textbf{PP}              &   \textbf{0s}    &   \textbf{0.5s}   &   \textbf{1s}     &   \textbf{1.5s}   &   \textbf{2s}     &   \textbf{2.5s}\\ \hline\hline
    \textbf{0s}              &   \blackcell     &   \greencell      &                   &   0.021           &   0.002           &                \\ \hline
    \textbf{0.5s}            &   \greencell     &   \blackcell      &   0.034           &   0.02            &   0.033           &   0.033        \\ \hline
    \textbf{1s}              &                  &   0.034           &   \blackcell      &   \redcell        &   \redcell        &   \redcell     \\ \hline
    \textbf{1.5s}            &   0.021          &   0.02            &   \redcell        &   \blackcell      &   \redcell        &   \redcell     \\ \hline
    \textbf{2s}              &   0.002          &   0.033           &   \redcell        &   \redcell        &   \blackcell      &   \redcell     \\ \hline
    \textbf{2.5s}            &                  &   0.033           &   \redcell        &   \redcell        &   \redcell        &   \blackcell   \\ \hline
  \end{tabular}
  \label{tab-loi:sm-perceivedperformance}
  \renewcommand{\arraystretch}{1}
\end{table}

\begin{table}[t!]
  \centering
  \caption{Significance matrix for differences in effort between levels of information loss. The shaded regions indicate the two ranges of usability. The cell entries are the $p$-values based on the Friedman test. The empty cells represent a comparison with no significant difference.}
  \renewcommand{\arraystretch}{1.1}
  \begin{tabular}{c||C{1.2cm}|C{1.2cm}|C{1.2cm}|C{1.2cm}|C{1.2cm}|C{1.2cm}}
    \hline\hline
    \textbf{E}              &   \textbf{0s}    &   \textbf{0.5s}   &   \textbf{1s}     &   \textbf{1.5s}   &   \textbf{2s}     &   \textbf{2.5s}\\ \hline\hline
    \textbf{0s}              &   \blackcell     &   \greencell      &  0.005            &                   &   0.018           &   0.039        \\ \hline
    \textbf{0.5s}            &   \greencell     &   \blackcell      &   0.029           &                   &   0.013           &   0.046        \\ \hline
    \textbf{1s}              &   0.005          &   0.029           &   \blackcell      &   \redcell        &   \redcell        &   \redcell     \\ \hline
    \textbf{1.5s}            &                  &                   &   \redcell        &   \blackcell      &   \redcell        &   \redcell     \\ \hline
    \textbf{2s}              &   0.018          &   0.013           &   \redcell        &   \redcell        &   \blackcell      &   \redcell     \\ \hline
    \textbf{2.5s}            &   0.039          &   0.046           &   \redcell        &   \redcell        &   \redcell        &   \blackcell   \\ \hline
  \end{tabular}
  \label{tab-loi:sm-effort}
  \renewcommand{\arraystretch}{1}
\end{table}

\begin{table}[t!]
  \centering
  \caption{Significance matrix for differences in frustration between levels of information loss. The shaded regions indicate the two ranges of usability. The cell entries are the $p$-values based on the Friedman test. The empty cells represent a comparison with no significant difference.}
  \renewcommand{\arraystretch}{1}
  \begin{tabular}{c||C{1.2cm}|C{1.2cm}|C{1.2cm}|C{1.2cm}|C{1.2cm}|C{1.2cm}}
    \hline\hline
    \textbf{F}              &   \textbf{0s}    &   \textbf{0.5s}   &   \textbf{1s}     &   \textbf{1.5s}   &   \textbf{2s}     &   \textbf{2.5s}\\ \hline\hline
    \textbf{0s}              &   \greencell     &                   &  0.012            &  0.002            &  $<$0.001           &  $<$0.001        \\ \hline
    \textbf{0.5s}            &   0.012          &   \blackcell      &  $<$0.001           &  0.02             &  $<$0.001           &   0.001        \\ \hline
    \textbf{1s}              &   0.002          &   $<$0.001          &   \blackcell      &                   &                   &   0.018        \\ \hline
    \textbf{1.5s}            &   0.001          &   0.02            &                   &   \blackcell      &   0.029           &   0.005        \\ \hline
    \textbf{2s}              &   $<$0.001         &   $<$0.001          &                   &   0.029           &   \blackcell      &     \redcell    \\ \hline
    \textbf{2.5s}            &   $<$0.001         &   0.001           &   0.018           &   0.005           &   \redcell        &   \blackcell   \\ \hline
  \end{tabular}
  \label{tab-loi:sm-frustration}
  \renewcommand{\arraystretch}{1}
\end{table}

\begin{table}[t!]
  \centering
  \caption{Significance matrix for differences in visual discomfort between levels of information loss. The shaded regions indicate the two ranges of usability. The cell entries are the $p$-values based on the Friedman test. The empty cells represent a comparison with no significant difference.}
  \renewcommand{\arraystretch}{1}
  \begin{tabular}{c||C{1.2cm}|C{1.2cm}|C{1.2cm}|C{1.2cm}|C{1.2cm}|C{1.2cm}}
    \hline\hline
    \textbf{VD}              &   \textbf{0s}    &   \textbf{0.5s}   &   \textbf{1s}     &   \textbf{1.5s}   &   \textbf{2s}     &   \textbf{2.5s}\\ \hline\hline
    \textbf{0s}              &   \blackcell     &   \greencell      &   \greencell      &   0.035           &   0.001           &    0.013       \\ \hline
    \textbf{0.5s}            &   \greencell     &   \blackcell      &   \greencell      &                   &   0.004           &    0.021       \\ \hline
    \textbf{1s}              &   \greencell     &   \greencell      &   \blackcell      &   \redcell        &   \redcell        &    \redcell    \\ \hline
    \textbf{1.5s}            &   0.035          &                   &   \redcell        &   \blackcell      &   \redcell        &    \redcell    \\ \hline
    \textbf{2s}              &   0.001          &   0.004           &   \redcell        &   \redcell        &   \blackcell      &    \redcell    \\ \hline
    \textbf{2.5s}            &   0.013          &   0.021           &   \redcell        &   \redcell        &   \redcell        &   \blackcell   \\ \hline
  \end{tabular}
  \label{tab-loi:sm-visual}
  \renewcommand{\arraystretch}{1}
\end{table}

\begin{table}[t!]
  \centering
  \caption{Overlaid significance matrices for determining the range of operability.}
  \renewcommand{\arraystretch}{1}
  \begin{tabular}{c||C{1.2cm}|C{1.2cm}|C{1.2cm}|C{1.2cm}|C{1.2cm}|C{1.2cm}}
    \hline\hline
    \textbf{VD}              &   \textbf{0s}    &   \textbf{0.5s}   &   \textbf{1s}     &   \textbf{1.5s}   &   \textbf{2s}     &   \textbf{2.5s}\\ \hline\hline
    \textbf{0s}              &   \dddgreencell  &   \ddgreencell    &   \dgreencell     &                   &                   &                \\ \hline
    \textbf{0.5s}            &   \ddgreencell   &   \ddgreencell    &   \dgreencell     &                   &                   &                \\ \hline
    \textbf{1s}              &   \dgreencell    &   \dgreencell     &   \greenredcell   &   \dredcell       &   \dredcell       &    \dredcell    \\ \hline
    \textbf{1.5s}            &                  &                   &   \dredcell       &   \dredcell       &   \dredcell       &    \dredcell    \\ \hline
    \textbf{2s}              &                  &                   &   \dredcell       &   \dredcell       &   \ddredcell      &    \ddredcell    \\ \hline
    \textbf{2.5s}            &                  &                   &   \dredcell       &   \dredcell       &   \ddredcell      &   \dddredcell   \\ \hline
  \end{tabular}
  \label{tab-loi:sm-overlaid}
  \renewcommand{\arraystretch}{1}
\end{table}

\subsection{Information Loss Main Study}
\label{sec:mainstudy}

\textbf{Experimental Setup.}
The final study we performed concerns the role of information loss in remote interaction between multiple humans and multiple robots. A particular aspect we intent to explore is the role of heterogeneous information loss across operators. To this aim, we consider also the homogeneous case as a baseline. From the results of the pilot study in Sec.~\ref{sec:pilotstudy}, we identified two levels of information loss: a low level, corresponding to high usability (\unit[0.25]{s}), and a high level, corresponding to low usability (\unit[2.25]{s}). We again used our collective transport game scenario and asked every participant to perform four experiments, one for each combination of levels of information loss for the operators. Once more, we randomized the order of the tasks to mitigate learning effects and other artifacts. In the following figures and tables, we use the following symbols to denote the four cases:
\begin{itemize}
\item Ho$_{LL}$: low homogeneous information loss;
\item Ho$_{HH}$: high homogeneous information loss;
\item He$_{LH}$: heterogeneous information loss in which operator 1 has low loss and operator 2 has high loss;
\item He$_{HL}$: heterogeneous information loss in which the operators are reversed with respect to He$_{LH}$.
\end{itemize}

\textbf{Hypotheses.}
We seek to validate the following working hypotheses:
\begin{itemize}
\item \textbf{H$_{IL}$1}: The case of low homogeneous information loss is the best overall with respect to the other cases in terms of measured metrics.
\item \textbf{H$_{IL}$2}: The operators prefer low homogeneous information loss to the other cases.
\item \textbf{H$_{IL}$3}: In the heterogeneous information loss case, operators prefer to be the ones with low information loss.
\item \textbf{H$_{IL}$4}: Operators prefer to experience high information loss in the heterogeneous case to being in the high homogeneous loss case.
\end{itemize}

\textbf{Participant Sample.}
We randomly paired the participants of the pilot study, forming 10 teams. Each team went through the four aforementioned cases.

\textbf{Procedures.}
Each session took approximately 105 minutes. Each session began with a training period, followed by 12 minutes of independent exploration of the system by the participants. After each session, each participant had to answer a subjective questionnaire.

\textbf{Metrics.}
We recorded subjective objective metrics for each participant and for each case. We used the same metrics presented in Sec.~\ref{sec-loi:tcstudy}. In addition, we recorded the number of interactions the participants made with the interface, as well as the time interval between those interactions. This allowed us to analyze the difference in workload between operators of the same team.

\begin{table}[t!]
  \centering
  \caption{Results of subjective scales with relationships between levels of information loss. The relationships are based on mean ranks obtained through Friedman tests. The symbol $^*$ denotes a significant difference ($p<0.05$) and the symbol $^{**}$ denotes a marginally significant difference ($p<0.10$). The symbol $^-$ denotes negative scales where lower ranking is better.}
  \renewcommand{\arraystretch}{1}
  \begin{tabular}{c|c|c|c}
    \hline
    \textbf{Attributes}            & \textbf{Relationship}                  & \textbf{$\chi^2(3)$}   & \textbf{$p$-value}    \\ \hline
    \multicolumn{4}{c}{\textbf{SART SUBJECTIVE SCALE}}                                                                       \\ \hline\hline
    Instability of Situation$^-$   & Ho$_{HH}>$He$_{HL}>$He$_{LH}>$Ho$_{LL}$& $21.924$                & $<0.001$             \\
    Complexity of Situation$^-$    & Ho$_{HH}>$He$_{HL}>$He$_{LH}>$Ho$_{LL}$& $26.024$                & $<0.001$             \\
    Variability of Situation$^-$   & Ho$_{HH}>$He$_{HL}>$He$_{LH}>$Ho$_{LL}$& $27.862$                & $<0.001$             \\
    Arousal                        & Ho$_{LL}>$He$_{LH}>$Ho$_{HH}>$He$_{HL}$& $18.850$                & $<0.001$             \\
    Concentration of Attention     & Ho$_{LL}>$He$_{LH}>$He$_{HL}>$Ho$_{HH}$& $16.088$                & $<0.001$             \\
    Spare Mental Capacity          & Ho$_{LL}>$He$_{LH}>$Ho$_{HH}>$He$_{HL}$& $10.112$                & $0.018$              \\
    Information Quantity           & Ho$_{LL}>$He$_{LH}>$Ho$_{HH}>$He$_{HL}$& $7.014$                 & $0.071$              \\
    Information Quality            & Ho$_{LL}>$He$_{LH}>$He$_{HL}>$Ho$_{HH}$& $11.464$                & $0.009$               \\
    Familiarity with Situation     & Ho$_{LL}>$He$_{LH}>$Ho$_{HH}>$He$_{HL}$& $6.949$                 & $0.074$              \\ \hline\hline
    \multicolumn{4}{c}{\textbf{NASA TLX SUBJECTIVE SCALE}}                                                                   \\ \hline\hline
    Mental Demand$^-$              & Ho$_{HH}>$He$_{HL}>$He$_{LH}>$Ho$_{LL}$& $15.112$                & $0.02$               \\
    Physical Demand$^-$            & Ho$_{HH}$=He$_{HL}>$He$_{LH}>$Ho$_{LL}$& $9.089$                 & $0.028$              \\
    Temporal Demand$^-$            & not significant                        & $5.447$                 & $0.142$              \\
    Performance                    & Ho$_{LL}>$He$_{LH}>$He$_{HL}>$Ho$_{HH}$& $37.893$                & $<0.001$             \\
    Effort$^-$                     & Ho$_{HH}>$He$_{HL}>$He$_{LH}>$Ho$_{LL}$& $23.053$                & $<0.001$             \\
    Frustration$^-$                & Ho$_{HH}>$He$_{HL}>$He$_{LH}>$Ho$_{LL}$& $21.124$                & $<0.001$             \\ \hline\hline
    \multicolumn{4}{c}{\textbf{TRUST SUBJECTIVE SCALE}}                                                                      \\ \hline\hline
    Competence                     & Ho$_{LL}>$He$_{LH}>$He$_{HL}>$Ho$_{HH}$& $31.461$                & $<0.001$             \\
    Predictability                 & Ho$_{LL}>$He$_{LH}>$He$_{HL}>$Ho$_{HH}$& $31.644$                & $<0.001$             \\
    Reliability                    & Ho$_{LL}>$He$_{LH}>$He$_{HL}>$Ho$_{HH}$& $33.737$                & $<0.001$             \\
    Faith                          & Ho$_{LL}>$He$_{LH}>$He$_{HL}>$Ho$_{HH}$& $31.210$                & $<0.001$             \\
    Overall Trust                  & Ho$_{LL}>$He$_{LH}>$He$_{HL}>$Ho$_{HH}$& $35.083$                & $<0.001$             \\
    Accuracy                       & Ho$_{LL}>$He$_{LH}>$He$_{HL}>$Ho$_{HH}$& $29.254$                & $<0.001$             \\ \hline\hline
    \multicolumn{4}{c}{\textbf{INTERACTION SUBJECTIVE SCALE}}                                                                \\ \hline\hline
    Teammate's Intent              & not significant                        & $5.880$                 & $0.118$              \\
    Teammate's Action              & Ho$_{LL}>$Ho$_{HH}>$He$_{HL}>$He$_{LH}$& $7.718$                 & $0.052$              \\
    Task Progress                  & Ho$_{LL}>$He$_{LH}>$Ho$_{HH}>$He$_{HL}$& $18.854$                & $<0.001$             \\
    Robot Status                   & Ho$_{LL}>$He$_{LH}>$Ho$_{HH}>$He$_{HL}$& $34.420$                & $<0.001$             \\
    Information Clarity            & Ho$_{LL}>$Ho$_{HH}>$He$_{LH}>$He$_{HL}$& $6.703$                 & $0.082$              \\ \hline
  \end{tabular}
  \label{tab-loi:results-subjective}
  \renewcommand{\arraystretch}{1}
\end{table}

\begin{table}[t!]
  \centering
  \caption{Results of objective metrics with relationships between levels of information loss. The relationships are based on mean ranks obtained through Friedman tests. The symbol $^*$ denotes a significant difference ($p<0.05$) and the symbol $^{**}$ denotes a marginally significant difference ($p<0.10$). The symbol $^-$ denotes negative scales where lower ranking is better.}
  \renewcommand{\arraystretch}{1}
  \begin{tabular}{c|c|c|c}
    \hline
    \textbf{Attributes}            & \textbf{Relationship}                  & \textbf{$\chi^2(3)$}   & \textbf{$p$-value}    \\ \hline
    \multicolumn{4}{c}{\textbf{PERFORMANCE OBJECTIVE SCALE}}                                                     \\ \hline\hline
    Time Taken for the task        & Ho$_{HH}>$He$_{HL}$=He$_{LH}>$Ho$_{LL}$& $11.803$                & $0.008$              \\
    Number of Interactions         & He$_{LH}>$Ho$_{HH}>$Ho$_{LL}>$He$_{HL}$& $17.258$                & $0.008$              \\
    Time gap between interactions  & Ho$_{HH}>$He$_{HL}>$He$_{LH}>$Ho$_{LL}$& $11.220$                & $0.011$              \\ \hline
  \end{tabular}
  \label{tab-loi:results-quantitative}
  \renewcommand{\arraystretch}{1}
\end{table}

\begin{table}[h]
  \centering
  \caption{Ranking scores based on the Borda count. The gray cells indicate the best case for each type of ranking.}
  \renewcommand{\arraystretch}{1}
  \begin{tabular}{c|c|c|c|c}
    \hline\hline
    Borda Count                                                                 & Ho$_{LL}$         & Ho$_{HH}$    & He$_{LH}$    & He$_{HL}$   \\ \hline\hline
    Based on Collected Data Ranking (Tables~\ref{tab-loi:results-subjective} \&~\ref{tab-loi:results-quantitative})   & \greycell{104}    & 36.5         & 74.5         & 45          \\
    Based on Preference Data Ranking (Fig.~\ref{fig-loi:task_usability})    & \greycell{77}     & 29           & 52           & 42          \\ \hline
  \end{tabular}
  \label{tab-loi:borda}
  \renewcommand{\arraystretch}{1}
\end{table}

\begin{table}[h!]
  \centering
  \caption{Results of subjective scales with attribute comparison between operators of the same team. The comparisons are based on mean ranks obtained through the Friedman test. The grey cells represent significant differences between operators in the same team. }
  \renewcommand{\arraystretch}{1}
  \begin{tabular}{c||c|c||c|c||c|c}
    \hline
    \multirow{3}{*}{\textbf{Attributes}}& \multicolumn{4}{c||}{\textbf{Homogeneous IL}}                                 & \multicolumn{2}{c}{\multirow{2}{*}{\begin{tabular}[c]{@{}c@{}}\textbf{Heterogeneous}\\\textbf{IL}\end{tabular}}}\\ \cline{2-5}
    & \multicolumn{2}{c||}{\textbf{Ho$_{LL}$}}  & \multicolumn{2}{c||}{\textbf{Ho$_{HH}$}}& \multicolumn{2}{l}{}                     \\ \cline{2-7}
    & \textbf{$\chi^2$} & \textbf{$p$-value}&\textbf{$\chi^2$} & \textbf{$p$-value} & \textbf{$\chi^2$} & \textbf{$p$-value}   \\ \hline
    \multicolumn{7}{c}{\textbf{SART SUBJECTIVE SCALE}}                                                                       \\ \hline\hline
    Instability of Situation       & 0                    & 1                 & 0                   & 1                  & 0.6                  & 0.439                \\
    Complexity of Situation        & \greycell{3}         & \greycell{0.083}  & 0                   & 1                  & 0.529                & 0.467                \\
    Variability of Situation       & 2.667                & 0.102             & 1.286               & 0.257              & 1.143                & 0.285                \\
    Arousal                        & 0.5                  & 0.480             & 0.667               & 0.414              & \greycell{7.143}     & \greycell{0.008}     \\
    Concentration of Attention     & 2.667                & 0.102             & 0.667               & 0.414              & \greycell{2.778}     & \greycell{0.096}     \\
    Spare Mental Capacity          & 0.2                  & 0.655             & 1.286               & 0.257              & \greycell{5.444}     & \greycell{0.02}      \\
    Information Quantity           & 0.5                  & 0.480             & 0.5                 & 0.480              & 1.667                & 0.197                \\
    Information Quality            & 0.5                  & 0.480             &  0.143              & 0.750              & \greycell{5.444}     & \greycell{0.02}      \\
    Familiarity with Situation     & 0.2                  & 0.655             & 0                   & 1                  & 0.057                & 0.796                \\\hline\hline
    \multicolumn{7}{c}{\textbf{NASA TLX SUBJECTIVE SCALE}}                                                                   \\ \hline\hline
    Mental Demand                  & 0                    & 1                 & 0.5                 & 0.48               & \greycell{3.257}     & \greycell{0.071}     \\
    Physical Demand                & 0.333                & 0.564             & 0.111               & 0.739              & 1.143                & 0.285                \\
    Temporal Demand                & 1                    & 0.317             & 0.143               & 0.705              & 0.077                & 0.782                \\
    Performance                    & 2                    & 0.157             & 0.111               & 0.739              & \greycell{7.143}     & \greycell{0.008}     \\
    Effort                         & 0                    & 1                 & 0.2                 & 0.655              & \greycell{5.444}     & \greycell{0.02}      \\
    Frustration                    & 0.333                & 0.564             & 0                   & 1                  & \greycell{3.267}     & \greycell{0.071}     \\\hline\hline
    \multicolumn{7}{c}{\textbf{TRUST SUBJECTIVE SCALE}}                                                                      \\ \hline\hline
    Competence                     & 0                    & 1                 & 1.8                 & 0.180              & \greycell{9.308}     & \greycell{0.002}     \\
    Predictability                 & 2                    & 0.157             & 0                   & 1                  & \greycell{6.231}     & \greycell{0.013}     \\
    Reliability                    & 0.333                & 0.564             & 2.667               & 0.102              & \greycell{6.231}     & \greycell{0.013}     \\
    Faith                          & 0.333                & 0.564             & 0.667               & 0.414              & \greycell{3.769}     & \greycell{0.052}     \\
    Overall Trust                  & 0                    & 1                 & 0.2                 & 0.655              & \greycell{6.231}     & \greycell{0.013}     \\
    Accuracy                       & 0                    & 1                 & 0.2                 & 0.655              & \greycell{5.444}     & \greycell{0.02}      \\\hline\hline
    \multicolumn{7}{c}{\textbf{INTERACTION SUBJECTIVE SCALE}}                                                                \\ \hline\hline
    Teammate's Intent              & 0.667                & 0.414             & 0                   & 1                  & 0.057                & 0.795                \\
    Teammate's Action              & 1.8                  & 0.180             & 0.143               & 0.705              & 0                    & 1                    \\
    Task Progress                  & 0.333                & 0.564             & 2.667               & 0.102              & 2.579                & 0.108                \\
    Robot Status                   & 0.667                & 0.414             & 0.4                 & 0.527              & \greycell{5.333}     & \greycell{0.021}     \\
    Information Clarity            & 0.143                & 0.705             & 0.5                 & 0.480              & 0.286                & 0.593                \\\hline
  \end{tabular}
  \label{tab-loi:results-ootl-subjective}
  \renewcommand{\arraystretch}{1}
\end{table}

\begin{table}[t!]
  \centering
  \caption{Results of quantitative scales with attribute comparison between operators of the same team. The comparisons are based on mean ranks obtained through the Friedman test. The grey cells represent significant differences between operators in the same team. }
  \renewcommand{\arraystretch}{1}
  \begin{tabular}{c||c|c||c|c||c|c}
    \hline
    \multirow{3}{*}{\textbf{Attributes}}& \multicolumn{4}{c||}{\textbf{Homogeneous IL}}                                 & \multicolumn{2}{c}{\multirow{2}{*}{\begin{tabular}[c]{@{}c@{}}\textbf{Heterogeneous}\\\textbf{IL}\end{tabular}}}\\ \cline{2-5}
    & \multicolumn{2}{c||}{\textbf{Ho$_{LL}$}}  & \multicolumn{2}{c||}{\textbf{Ho$_{HH}$}}& \multicolumn{2}{l}{}                     \\ \cline{2-7}
    & \textbf{$\chi^2$} & \textbf{$p$-value}&\textbf{$\chi^2$} & \textbf{$p$-value} & \textbf{$\chi^2$} & \textbf{$p$-value}   \\ \hline
    \multicolumn{7}{c}{\textbf{PERFORMANCE OBJECTIVE SCALE}}                                                     \\ \hline\hline
    Number of Interactions         & 0.111                & 0.739             & 1                   & 0.317              & 0                    & 1                    \\
    Time gap between interactions  & 0.4                  & 0.527             & 1.6                 & 0.206              & 0                    & 1                    \\\hline
  \end{tabular}
  \label{tab-loi:results-ootl-quantitative}
  \renewcommand{\arraystretch}{1}
\end{table}

\begin{figure}[t]
  \centering
  \includegraphics[width=0.72\textwidth, trim={0 0 0 1cm},clip]{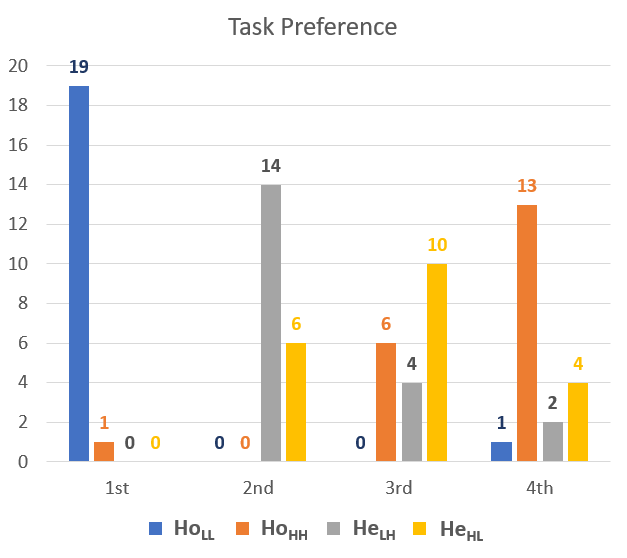}
  \caption{Operator preferences in information loss.}
  \label{fig-loi:task_usability}
\end{figure}

\textbf{Results.}
Tables~\ref{tab-loi:results-subjective} and~\ref{tab-loi:results-quantitative} show the summarized results for the subjective scales and the objective metrics. We used the Friedman test to establish significance between different cases. We formed rankings based on the mean ranks for all the attributes that showed statistical significance ($p<0.05$) or marginal significance ($p<0.10$). Tables~\ref{tab-loi:results-ootl-subjective} and~\ref{tab-loi:results-ootl-quantitative} report an imbalance in awareness, workload, trust and interaction quality between operators of the same team in tasks with heterogeneous information loss. Fig.~\ref{fig-loi:task_usability} shows which information loss cases were preferred by each operator. We used the Borda count~\cite{black1976partial} to calculate the overall ranking. Table~\ref{tab-loi:borda} shows the results of the Borda count for each category.

\subsubsection{Pilot and Main Study: Comparative Analysis}
\label{sec-loi:discussion}

\begin{figure}[t]
  \centering
  \includegraphics[width=0.72\textwidth]{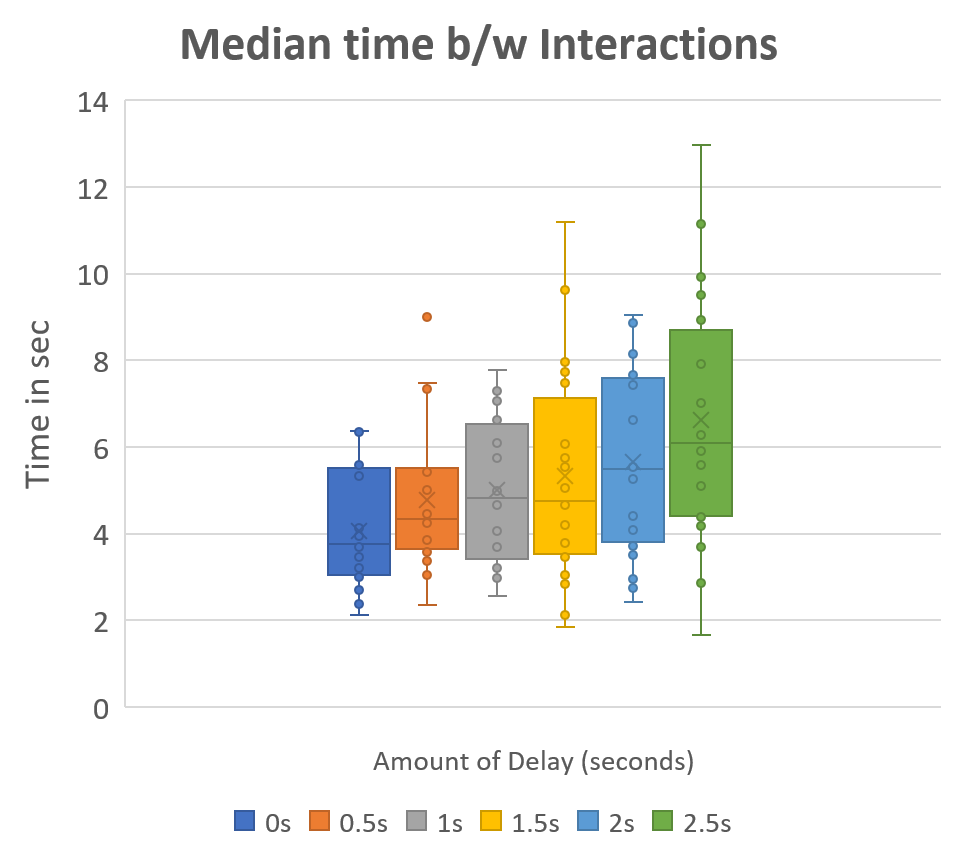}
  \caption{Box plot of the recorded time gap between each interaction for each operator.}
  \label{fig-loi:timegap}
\end{figure}

\textbf{Pilot Study Data Analysis.}
Tables~\ref{tab-loi:sm-performance}-\ref{fig-loi:vd} and Figures~\ref{fig-loi:performance}-\ref{fig-loi:frustration} indicate that, with the increase in information loss, the workload experienced by the operator increases while performance degrades. We compared the number of interactions made with each level of information loss, and found no significant difference. We also recorded the time interval between interactions. The box plot of the median values (shown in Fig.~\ref{fig-loi:timegap}) indicates a significant increase ($\chi^2(1) = 30.486$, $p < 0.001$) in time waited between interactions, according to the well-known \emph{waiting strategy} observed in user studies with traditional tele-operation and remote interaction systems~\cite{ellis2004generalizeability}.

\textbf{Pilot Study Behavioral Analysis.}
We observed the behaviour of the operators during and after each session. Two operators (out of 20) chose to stop their session with \unit[2]{s} and \unit[2.5]{s} of information loss. They reported that they had reached their ability to handle the high information loss. Eleven operators reported that they had reached their limit of frustration at \unit[2.5]{s}, but nevertheless chose to continue because of their \emph{never give up} attitude and their willingness to help our research. Seven operators reported that they could have handled higher than \unit[2.5]{s} of loss because of their past experience with laggy systems and internet. As for the discomfort experienced by the operators, three operators started experiencing discomfort with \unit[1]{s} of information loss; four operators with values over \unit[1.5]{s}; three operators with information loss over \unit[2]{s}; and six operators with information loss over \unit[2.5]{s}. The reported discomfort included a slight headache and fatigue in their eyes. As a part of the exit interview for the pilot study, we asked the participants if the task order assigned to them impacted their performance in the study. The participants in the increasing order of information loss reported that the increase in loss made them ready for the next task and they expected the loss to increase. They reported that, with each task, the familiarity with experiencing loss was increasing, causing them to be better trained at handling it. All the participants in this category reported that they would have been more frustrated if the task ordering was reversed and they would be most frustrated if they had to experience the maximum information loss in the first task. However, the participants in the decreasing order of information loss reported that they would have been more frustrated if the information loss were increasing in each task.  All the participants in this cohort reported that, as the loss was decreasing, they knew the worst was over and the tasks will only get easier from there on. We call this the \emph{count one's blessing} phenomenon: the participants preferred and defended their task order, assuming that the reverse order would only harm their performance and interaction quality.

\textbf{Main Study Data Analysis.} Table~\ref{tab-loi:borda} shows that Ho$_{LL}$ is the best information loss case both in terms of usability preference and according to the data collected during the user study. This supports our hypotheses \textbf{H$_{IL}$1} and \textbf{H$_{IL}$2} that low homogeneous information loss is the best overall case. The He$_{LH}$ case is the next best choice for the participants, indicating preference for low personal information loss. This supports hypothesis, \textbf{H$_{IL}$3}. The He$_{HL}$ case is the third choice, showing that either operator experiencing low loss is still better than both operators experiencing high information loss. This supports hypothesis \textbf{H$_{IL}$4}.

\textbf{Main Study Behavioral Analysis.} We also observed the behaviour of the operators during and after the sessions. Based on the preference shown in Fig.~\ref{fig-loi:task_usability}, we could categorize the participants in four typologies. (a) \emph{The Egocentrics}: ten participants gave higher preference to the tasks with low information loss, and lower preference to the tasks with high information loss. However, when they had to rank their preference between the options of choosing low and high information loss for themselves and give the other to their teammate, the participants opted for low information loss even though that meant that their teammate might get more frustrated by experiencing higher loss. (b) \emph{The Altruists}: five participants preferred to handle high information loss so that their teammate might face lower levels of frustration while interacting with a low information loss. These participants, the altruists, reported that they were confident in their ability to handle high information loss, and with their teammate experiencing low information loss their chances of completing the task might increase. (c) \textit{The Egalitarians}: four participants preferred homogeneous loss over heterogeneous loss, even if that means that both operators would have to experience a high information loss. These participants reported that, with homogeneous information loss, they could actively interact with their teammate and handle equal workload, which they did not experience in tasks with heterogeneous information loss. (d) \textit{The Thinker}: one participant preferred high information loss over low information loss. This participant reported that high information loss provided more time to think before making the next step and could interact more with the fellow teammate while doing so.

\textbf{On the Out-of-the-Loop Performance Problem.}
Tables~\ref{tab-loi:results-ootl-subjective} and~\ref{tab-loi:results-ootl-quantitative} show that the participants experienced unbalanced awareness, workload, trust and interaction quality, while engaging in the tasks with heterogeneous information loss. This imbalance indicates that the operator experiencing high information loss will go \textit{out of the loop}~\cite{endsley1995out,gouraud2017autopilot}. However, the interaction quality scales show that the significant difference in information awareness is observed only for the robot-level information and not on operator-level information. We conclude this as there was no loss or delay experienced in the communication channel for this user study; future work could investigate the impact of loss of communication between the operators.

\section{Conclusions and Future Work}
\label{sec:conclusion}

In this paper, we studied the effects of transparency, inter-human communication, and information loss on multi-human multi-robot interaction. We first performed a study of the most effective interface elements to support information transparency and inter-operator transparency. We analyzed the usability of our interface through a user study with 28 operators measuring awareness, workload, trust, and interaction efficiency. The findings of the user study indicated mixed transparency as the best transparency mode and mixed communication as the best communication mode.

We then studied the effects of information loss on the performance of the operators. We performed two user studies. The first, a pilot study, aimed to identify the amount of information loss that can be considered noticeable but bearable for the average operator, and which amount of information loss is unbearable. Using the result of this study, we performed a thorough exploration of the role of information loss in multi-operator scenarios, comparing heterogeneous and homogeneous cases. We derived a set of behavioral typologies of users, revealing that remote interaction must consider personal preferences and individual attitude when forming groups of operators.

Future work will focus on the role of training in multi-human multi-robot interaction. In this study, we assumed that no participant had prior experience with the interface, and we provided minimal guidance to avoid biasing our studies. However, effective multi-human multi-robot interaction for complex missions cannot ignore the need for training and proper teaming according to individual skills.

\bibliographystyle{spmpsci}
\bibliography{ref}

\end{document}